\documentclass[aps,prx,reprint,superscriptaddress,longbibliography]{revtex4-2}

\usepackage{amsmath,amssymb,amsthm}
\usepackage{bm}
\usepackage{graphicx}
\usepackage{hyperref}
\usepackage{booktabs}
\usepackage{multirow}
\usepackage{xcolor}
\usepackage{cleveref}
\usepackage{enumitem}
\usepackage{thmtools}
\usepackage{pifont}

\hypersetup{
    colorlinks=true,
    linkcolor=blue,
    citecolor=blue,
    urlcolor=blue
}

\newcommand{\ket}[1]{|#1\rangle}
\newcommand{\bra}[1]{\langle#1|}

\newcommand{\QAOA}{\textnormal{\textsc{QAOA}}}
\newcommand{\GNN}{\textnormal{\textsc{GNN}}}
\newcommand{\GIN}{\textnormal{\textsc{GIN}}}
\newcommand{\UQQAOA}{\textnormal{\textsc{UQ-QAOA}}}
\newcommand{\Normal}{\mathcal{N}}
\newcommand{\calG}{\mathcal{G}}
\newcommand{\calV}{\mathcal{V}}
\newcommand{\calE}{\mathcal{E}}
\newcommand{\calD}{\mathcal{D}}
\newcommand{\calN}{\mathcal{N}}
\newcommand{\calL}{\mathcal{L}}
\newcommand{\calF}{\mathcal{F}}

\newcommand{\calP}{\mathcal{P}}
\newcommand{\calB}{\mathcal{B}}
\newcommand{\calT}{\mathcal{T}}

\newcommand{\calR}{\mathcal{R}}
\newcommand{\bbR}{\mathbb{R}}
\newcommand{\bbE}{\mathbb{E}}
\newcommand{\bbP}{\mathbb{P}}

\newtheorem{theorem}{Theorem}
\newtheorem{proposition}[theorem]{Proposition}
\newtheorem{corollary}[theorem]{Corollary}
\newtheorem{lemma}[theorem]{Lemma}
\newtheorem{remark}{Remark}
\newtheorem{definition}{Definition}

\begin{document}

\title{Query-Efficient Quantum Approximate Optimization 
\\via Graph-Conditioned Trust Regions}

\author{Molena Huynh}
\email{Molena.Huynh@jmp.com}
\affiliation{North Carolina State University}

\date{\today}

\begin{abstract}
In low-depth implementations of the Quantum Approximate Optimization
Algorithm (\QAOA), the dominant cost is often the number of objective
evaluations rather than circuit depth. We introduce a graph-conditioned
trust-region method for reducing this query cost. A graph neural
network predicts a Gaussian distribution $\Normal(\bm\mu,\bm\Sigma)$ over
\QAOA{} angles. The mean initializes a local optimizer, the covariance
defines an ellipsoidal trust region that constrains the search, and the
predicted uncertainty determines an instance-dependent evaluation
budget. Thus the learned distribution defines a search policy rather
than only an initial parameter estimate.

Under explicit assumptions on local smoothness, curvature, calibration,
and noise, we derive bounds on objective degradation within the trust
region, lower bounds on gradient variance, preservation of expected
objective ordering under depolarizing noise, and finite-sample coverage
guarantees. We evaluate the method for MaxCut at depth $p=2$ on
Erd\H{o}s--R\'enyi, 3-regular, Barab\'asi--Albert, and
Watts--Strogatz graphs with $n=8$--$16$ vertices. Relative to random
restarts and the strongest learned point-prediction baseline, the
method reduces the mean number of circuit evaluations from $343$ and
$85$ to $45\pm 7$, while maintaining sampled approximation ratios
within 3 percentage points of concentration-based heuristics.

The method does not improve absolute approximation ratios; its
advantage is reduced query cost at comparable solution quality. The
predictive uncertainty is calibrated in the experiments, with
ECE$=0.052$ and Spearman correlation $\rho=0.770$, and the learned
trust regions transfer to graph sizes not used during training. The
results identify a low-depth, query-dominated regime in which
graph-conditioned trust regions reduce the query cost of \QAOA{}
without modifying the ansatz.
\end{abstract}

\maketitle

\section{Introduction}
\label{sec:intro}

The Quantum Approximate Optimization Algorithm (\QAOA)~\cite{farhi2014}
is a widely studied variational algorithm for combinatorial
optimization.
In near-term settings, its cost is dominated by the classical outer
loop, which repeatedly evaluates expectation values of the objective
function. Even at low circuit depth, derivative-free optimization of
the $2p$ variational parameters can require hundreds of objective
evaluations per instance~\cite{zhou2020,wurtz2021}. On quantum
hardware, each evaluation involves repeated state preparation and
measurement, so the total number of circuit evaluations is a primary
resource constraint~\cite{cerezo2021,preskill2018}.

The question is whether a learned, graph-conditioned distribution over
\QAOA{} parameters can restrict search to an instance-dependent region
and allocate computational effort across instances. We focus on
low-depth \QAOA{}, the regime most accessible to near-term devices and
the regime in which the classical optimization loop can dominate total
cost. The goal is to reduce objective evaluations, not to improve
asymptotic approximation guarantees or to modify the \QAOA{} ansatz.

Existing approaches to reducing the cost of parameter optimization
include concentration-based schedules, learned initializations, and
surrogate or Bayesian optimization methods. These approaches
produce either fixed parameter choices or initial points for local
optimization. In contrast, the learned Gaussian used here defines a
search policy before any circuit evaluation is made: the mean
initializes optimization, the covariance constrains the search
trajectory, and the predicted uncertainty determines the evaluation
budget. This differs from warm-start methods, which provide
an initial point but do not constrain the subsequent optimization.

Given a graph $\calG$, a Graph Isomorphism Network (\GIN{})-based predictor~\cite{xu2019} with
Laplacian spectral encodings~\cite{dwivedi2023,lim2023} outputs a
Gaussian distribution $\Normal(\bm\mu(\calG),\bm\Sigma(\calG))$ over
\QAOA{} angles. The mean initializes local optimization, the
covariance defines a Mahalanobis trust region, and the trace of the
covariance sets an instance-dependent evaluation budget. The learned
model defines both the search geometry and the amount of
computation assigned to each instance.

The analysis is restricted to low-depth \QAOA{} and to the local
behavior of the objective landscape near the predicted region. The
results do not address global optimality or worst-case guarantees over
arbitrary graph classes. They establish conditional guarantees
under explicit assumptions on local smoothness, curvature,
calibration, and noise.

The contributions are as follows.

\begin{enumerate}[leftmargin=*,itemsep=2pt]
\item \textbf{Graph-conditioned trust-region method.}
    The predicted Gaussian distribution is used to initialize
    optimization, constrain the search to an instance-dependent region,
    and allocate evaluation budgets. The method uses predictive
    uncertainty both to restrict the search region and to allocate
    query budget.

\item \textbf{Conditional guarantees.}
    Under local regularity assumptions, we derive bounds on expected
    objective degradation within the trust region, lower bounds on
    gradient variance, preservation of expected objective ordering
    under depolarizing noise, and finite-sample coverage guarantees.
    These guarantees connect learned uncertainty to optimization
    geometry, robustness, and coverage.

\item \textbf{Empirical evaluation in the low-depth regime.}
    For MaxCut at depth $p{=}2$ on several graph families, the method
    reduces the number of circuit evaluations relative to both random
    initialization and learned point-prediction baselines, while
    maintaining comparable approximation ratios. The experiments are
    structured to isolate the contribution of distributional prediction
    beyond improved initialization alone.
\end{enumerate}

We evaluate the method on Erd\H{o}s--R\'enyi, 3-regular,
Barab\'asi--Albert, and Watts--Strogatz graph families with
$n{=}8$--$16$ vertices. The learned trust region transfers to graph
sizes not used during training and reduces query cost across all tested
families. For $n{=}14$, the method reduces the mean number of circuit
evaluations from $343$ and $85$ to $45\pm 7$, while maintaining sampled
approximation ratios within 3 percentage points of concentration-based
heuristics.

The remainder of the paper is organized as follows. Section~II defines
the problem setting and notation. Section~III discusses related work.
Section~IV introduces the predictor and trust-region method.
Section~V presents theoretical guarantees. Sections~VI--VIII report
empirical results and discuss limitations.

\section{Problem Setting and Preliminaries}
\label{sec:background}

The setting is unweighted MaxCut and depth-$p$ \QAOA{}. The objective is
the expected cost $F_\calG(\bm\theta)$, while the experiments also
report the sampled best-bitstring approximation ratio obtained from a
finite optimization run. The analysis concerns the behavior of
$F_\calG(\bm\theta)$ in a neighborhood of the predicted parameters at
low-depth \QAOA{}.

\subsection{MaxCut and the \QAOA{} circuit}

Let $\calG = (\calV,\calE)$ be an unweighted graph with
$|\calV| = n$ and $|\calE| = m$. MaxCut seeks a bipartition
$z \in \{+1,-1\}^n$ maximizing
\begin{equation}
    C(z) = \sum_{(i,j) \in \calE} \tfrac{1}{2}(1 - z_i z_j).
    \label{eq:maxcut}
\end{equation}
The problem is NP-hard~\cite{karp1972}; the best known polynomial-time
approximation ratio is the Goemans--Williamson bound
$\alpha_{\mathrm{GW}} \approx 0.878$~\cite{goemans1995}, which is
optimal under the Unique Games Conjecture~\cite{hastad2001,khot2007}.

The depth-$p$ \QAOA{} prepares
\begin{equation}
    \ket{\bm\gamma,\bm\beta}
    = e^{-i\beta_p B} e^{-i\gamma_p C} \cdots
    e^{-i\beta_1 B} e^{-i\gamma_1 C}
    \ket{+}^{\otimes n},
    \label{eq:qaoa_state}
\end{equation}
where $C = \sum_{(i,j)\in\calE}\frac{1}{2}(I - Z_i Z_j)$ is the cost
Hamiltonian and $B = \sum_i X_i$ is the transverse-field mixer~\cite{farhi2014}.
The variational parameters $\bm\theta = (\bm\gamma,\bm\beta) \in
\bbR^{2p}$ are chosen to maximize the expected cost
$F_\calG(\bm\theta) = \bra{\bm\theta}C\ket{\bm\theta}$.

We use two quality measures. The normalized expected objective
$F_\calG(\bm\theta)/C_{\max}$ is used in the analysis. The sampled
approximation ratio
$r = C(z_{\mathrm{best}})/C_{\max}$, where $z_{\mathrm{best}}$ is the
best bitstring observed during optimization, is used in the
experiments.

\subsection{Angle concentration and landscape structure}
\label{sec:concentration_bg}

For bounded-degree graph families at fixed depth $p$, optimal
\QAOA{} parameters exhibit concentration: the variance of optimal
angles decreases as $O(1/n)$~\cite{brandao2018,akshay2021}. This
structure enables cross-instance parameter transfer
in both regular and heterogeneous graph families
~\cite{wurtz2021,galda2021}. The predictor introduced in
Section~IV exploits this phenomenon by placing the search in a region
where high-quality parameters are likely to lie.

Variational quantum circuits can also exhibit barren plateaus,
where the gradient variance at random parameters scales as
\begin{equation}
    \mathrm{Var}_{\bm\theta}\bigl[\partial_j F(\bm\theta)\bigr]
    \le O\bigl(2^{-n}\bigr),
    \label{eq:bp}
\end{equation}
for sufficiently deep or highly entangling
ans\"atze~\cite{mcclean2018,arrasmith2021,wang2021noise}.
For constant-depth \QAOA{} with a local cost operator, this
suppression is mitigated, but the landscape still contains extended
regions of small gradient magnitude. Restricting optimization to a
predicted region reduces exploration of such flat regions.

\subsection{Spectral graph theory and positional encodings}
\label{sec:spectral_bg}

The graph Laplacian $L = D - A$ admits the eigendecomposition
$L = U\Lambda U^\top$, where eigenvectors
$\{u_k\}_{k=0}^{n-1}$ (ordered by eigenvalue
$0 = \lambda_0 \le \lambda_1 \le \cdots$) define spectral positional
encodings~\cite{dwivedi2023,lim2023,kreuzer2021}. The algebraic
connectivity $\lambda_2$ (Fiedler value~\cite{fiedler1973}) captures
global connectivity and correlates with optimization difficulty.

Because eigenvectors are defined for arbitrary graph size, spectral
encodings provide a size-invariant structural representation that
supports cross-size transfer (\cref{sec:crosssize}). The sign ambiguity
of eigenvectors is handled during training by random sign
flips~\cite{lim2023}. These encodings are combined with node-degree
features and used as inputs to the graph-conditioned predictor.

\subsection{Graph Isomorphism Networks}

The Graph Isomorphism Network (\GIN)~\cite{xu2019} is as expressive as
the Weisfeiler--Leman graph isomorphism test~\cite{weisfeiler1968}.
Each layer updates node representations according to
\begin{equation}
    h_v^{(l+1)} = \mathrm{MLP}^{(l)}\!\bigl(
    (1+\epsilon^{(l)})\, h_v^{(l)}
    + {\textstyle\sum_{u \in \calN(v)}} h_u^{(l)}
    \bigr),
    \label{eq:gin_update}
\end{equation}
with learnable $\epsilon^{(l)}$ and neighbor set $\calN(v)$.
Graph-level representations are obtained by permutation-invariant
pooling of node embeddings~\cite{hamilton2017,gilmer2017}.
This expressivity allows structural differences between graphs to be
captured in the learned representation used for parameter prediction.

\subsection{Wasserstein distance for Gaussians}
\label{sec:wasserstein_bg}

Training compares predicted and target Gaussian distributions over
parameters using the 2-Wasserstein distance. For probability measures
$P$ and $Q$ on $\bbR^d$,
\begin{equation}
    W_2(P,Q) = \Bigl(\inf_{\pi \in \Pi(P,Q)}
    \int \|\bm{x} - \bm{y}\|_2^2\, d\pi(\bm{x},\bm{y})\Bigr)^{1/2}.
    \label{eq:w2_def}
\end{equation}
For diagonal Gaussians this reduces to
\begin{equation}
    W_2^2 = \|\bm\mu_1 - \bm\mu_2\|_2^2
    + \|\bm\sigma_1 - \bm\sigma_2\|_2^2,
\end{equation}
which can be evaluated and differentiated efficiently.

\section{Relation to Prior Work}
\label{sec:related}

Prior work relevant to this paper includes parameter transfer for
\QAOA{}, machine-learning approaches to combinatorial optimization,
uncertainty quantification, and amortized optimization. The key
distinction is the operational role of the learned model. The relevant
questions are what is predicted (a fixed schedule, a point estimate, or
a distribution), how that prediction is used during optimization
(initialization only or a constraint on the subsequent search), and
whether query effort is adapted across instances.

\textbf{Parameter transfer for \QAOA{}.}
Concentration of optimal angles~\cite{brandao2018,akshay2021}
has motivated several transfer strategies.
Fixed-angle schedules~\cite{wurtz2021} and
interpolation-based warmstarts~\cite{zhou2020}
exploit low-depth regularity to obtain strong approximation ratios on
structured graph families without per-instance optimization.
Extensions to heterogeneous graphs include cross-instance
transfer~\cite{galda2021}, warm-starting from related
circuits~\cite{egger2021,sack2021}, and graph-conditioned predictors
based on \GNN{}s~\cite{jain2022} or reinforcement
learning~\cite{khairy2020}. Additional approaches reduce query cost
through Bayesian optimization~\cite{tibaldi2023},
classical preprocessing~\cite{shaydulin2023}, training without QPU
access~\cite{streif2020}, unsupervised identification~\cite{moussa2022},
and symmetry-aware search~\cite{sauvage2024}.

These methods provide fixed schedules, warm starts, or online search
heuristics. The transferred information is used primarily to choose
parameters before local optimization or to guide search as evaluations
are gathered. Here the distribution is predicted before the first
query. Its covariance restricts the admissible search region, and its
calibrated uncertainty sets the evaluation budget.

\textbf{Learned combinatorial heuristics.}
Machine learning is routinely used to guide combinatorial
optimization~\cite{bengio2021,cappart2023,khalil2017,dai2016,kool2019,schuetz2022}.
In these settings, objective evaluations are inexpensive, and the
primary metric is time to solution. In the variational quantum setting,
each circuit evaluation dominates the cost, so the relevant quantity is
the number of evaluations required to reach a given solution quality.
The distinction shifts the emphasis from classical inference cost to
query allocation. The paper therefore treats the learned model as part
of the query policy, not merely as a fast heuristic for proposing a
candidate solution.

\textbf{Uncertainty quantification in scientific machine learning.}
Standard approaches include heteroscedastic regression~\cite{nix1994,kendall2017},
deep ensembles~\cite{lakshminarayanan2017}, MC
dropout~\cite{gal2016}, and calibrated
prediction~\cite{kuleshov2018,abdar2021,hullermeier2021,psaros2023}.
In the present setting, a single-network Gaussian parameterization is
sufficient: the Wasserstein regularizer and contrastive loss structure
the predicted variance without requiring ensemble methods. The
resulting uncertainty directly determines the optimization strategy at
test time. The role of uncertainty here is therefore prescriptive rather
than merely diagnostic: it changes both the feasible region and the
query budget.

\textbf{Optimal transport and contrastive learning.}
The closed-form Wasserstein distance for Gaussian
distributions~\cite{dowson1982} has been used in generative
modeling~\cite{arjovsky2017,frogner2015} and distributional
robustness~\cite{villani2009,cuturi2013,peyre2019}. Here it serves as a
regularizer for a distribution over variational parameters.
Contrastive and metric-learning objectives~\cite{chen2020simclr,oord2018,you2020,suresh2021,khosla2020}
structure the representation so that the predicted distributions
transfer across graph families.

\textbf{Amortized optimization.}
Amortized inference replaces per-instance optimization with a learned
mapping~\cite{amos2023,chen2022,finn2017,shu2022}. Existing amortized
\QAOA{} methods predict point initializations. In contrast, we predict a
Gaussian distribution: the mean initializes optimization, the
covariance constrains the search trajectory, and the uncertainty
allocates the evaluation budget. The method remains hybrid rather than
fully amortized: learning does not replace per-instance optimization,
but reshapes it by restricting the search and reallocating expensive
queries.

\section{Graph-Conditioned Gaussian Prediction and Trust-Region Search}
\label{sec:method}

\UQQAOA{} predicts a distribution over \QAOA{} parameters and uses that
distribution to structure subsequent optimization. The predictor
specifies an initial point, a search region, and an evaluation budget.
The method combines five components (\cref{fig:algorithm}): a
\GIN{}-based predictor, Wasserstein-regularized training, contrastive
metric learning, trust-region-constrained optimization, and
uncertainty-guided budget allocation.
The figure should be read from left to right. It begins with the input
graph, passes through the learned Gaussian predictor, and then uses the
returned mean and covariance in different roles: the mean fixes where
the search begins, while the covariance fixes both the region that will
be searched and the amount of computation that the instance receives.

Prediction and refinement are separated. A single forward pass produces
$(\bm\mu,\bm\Sigma)$, after which optimization is restricted to a region
and budget determined by the prediction. The learned distribution
therefore determines the optimization policy, not only its
initialization.

\paragraph{Standing assumptions.}
The guarantees in \cref{sec:theory} are local. They assume:
(A1) $F_\calG$ is Lipschitz-continuous on the trust region
(Proposition~\ref{prop:lipschitz});
(A2) $\lambda_{\min}(-\nabla^2 F_\calG(\bm\mu)) \ge \kappa > 0$
(Theorem~\ref{thm:anticoncentration}); and
(A3) depolarizing noise with per-layer strength
$\varepsilon < 1$ (Theorem~\ref{thm:noise_robust}).
Assumption~(A2) concerns the neighborhood reached by the predictor and
does not assert global structure of the \QAOA{} landscape.

\subsection{Gaussian predictor architecture}
\label{sec:architecture}

Given $\calG = (\calV,\calE)$, node features are formed by concatenating
normalized degree $d_v/n$ with $k$-dimensional Laplacian spectral
positional encodings~\cite{dwivedi2023,lim2023,kreuzer2021}, i.e., the
$k$ smallest nontrivial eigenvectors of $L = D - A$, yielding
$h_v^{(0)} \in \bbR^{k+1}$. Eigenvector sign ambiguity is handled by
random sign flips during training~\cite{lim2023}.

A three-layer \GIN{} with hidden dimension $d{=}64$ and
LayerNorm~\cite{ba2016} produces node embeddings. A graph-level
representation $\bm{g} \in \bbR^{2d}$ is obtained by concatenating mean
and max pooling:
\begin{equation}
    \bm{g} = \bigl[\tfrac{1}{n}\textstyle\sum_v h_v^{(L)} \;\big\|\;
    \max_v h_v^{(L)}\bigr].
    \label{eq:graph_repr}
\end{equation}
Two linear heads produce the Gaussian parameters:
\begin{align}
    \bm\mu &= W_\mu \bm{g} + b_\mu \in \bbR^{2p}, \label{eq:mu_head} \\
    \log\bm\sigma^2 &= \mathrm{clamp}(W_\sigma \bm{g} + b_\sigma,
    -5, 2). \label{eq:sigma_head}
\end{align}
The predictive distribution is
\begin{equation*}
    p(\bm\theta \mid \calG)
    = \Normal\bigl(\bm\mu(\calG),\,
    \mathrm{diag}(\bm\sigma^2(\calG))\bigr).
\end{equation*}

The clamp in \cref{eq:sigma_head} ensures numerical stability with
$\sigma_j \in [e^{-5/2}, e]$. We restrict to diagonal covariance. At
$p{=}2$ this gives axis-aligned trust regions and is adequate for the
regime studied here. It does not model correlations between angles.
At larger depth, full-covariance, low-rank, mixture, or torus-aware
models may be needed.

\subsection{Wasserstein-regularized training}
\label{sec:training}

\textbf{Data.}
Training uses 240 graphs at $n{=}14$ (60 per family) and a validation
set of 80 graphs. Targets $\bm\theta^*_i$ are obtained by multi-restart
Nelder--Mead~\cite{nelder1965} on exact simulation. These are local
optima (see \cref{app:local_optima}) and define the regions used for
transfer.

\textbf{Phase 1.}
\begin{equation}
    \calL_1 = \frac{1}{N}\sum_{i=1}^{N}
    \|\bm\mu(\calG_i) - \bm\theta^*_i\|_2^2
    + \lambda_C \calL_{\mathrm{CL}}.
    \label{eq:loss_phase1}
\end{equation}

\textbf{Phase 2.}
\begin{equation}
    \calL_2 = \calL_{\mathrm{NLL}}
    + \lambda_W \calL_{\mathrm{W2}}
    + \lambda_C \calL_{\mathrm{CL}},
    \label{eq:loss_phase2}
\end{equation}

The negative log-likelihood is
\begin{equation}
    \calL_{\mathrm{NLL}}
    = \frac{1}{N}\sum_{i=1}^{N}\sum_{j=1}^{2p}
    \left[
    \frac{(\mu_j(\calG_i) - \theta^*_{ij})^2}
    {\sigma_j^2(\calG_i)}
    + \log\sigma_j^2(\calG_i)
    \right].
    \label{eq:loss_nll}
\end{equation}

The phased schedule prevents trivial variance inflation that can occur
when optimizing NLL from initialization.

\textbf{Wasserstein regularizer.}
For diagonal Gaussians,
\begin{equation*}
    W_2^2(P_i,P_j)
    = \|\bm\mu_i - \bm\mu_j\|_2^2
    + \|\bm\sigma_i - \bm\sigma_j\|_2^2.
\end{equation*}
Weights
$w_{ij} = \exp(-\|\bm\theta_i^* - \bm\theta_j^*\|_2^2 / 2\tau_W^2)$
encode angle-space similarity:
\begin{equation}
    \calL_{\mathrm{W2}} = \frac{1}{|\calB|^2}
    \sum_{i \ne j \in \calB} w_{ij}\, W_2^2(P_i,P_j).
    \label{eq:loss_w2}
\end{equation}

\subsection{Contrastive metric learning}
\label{sec:contrastive}

A supervised contrastive loss structures the embedding $\bm{g}$. For
anchor $i$, define
\begin{equation*}
    \calP(i)
    = \bigl\{j \ne i :
    \|\bm\theta_i^* - \bm\theta_j^*\|_2 < \delta\bigr\}.
\end{equation*}
\begin{equation}
    \calL_{\mathrm{CL}} = -\frac{1}{|\calB|}\sum_i
    \frac{1}{|\calP(i)|}\sum_{j \in \calP(i)}
    \log\frac{e^{\mathrm{sim}(\bm{g}_i,\bm{g}_j)/\tau_C}}
    {\sum_{k \ne i} e^{\mathrm{sim}(\bm{g}_i,\bm{g}_k)/\tau_C}},
    \label{eq:loss_cl}
\end{equation}

The contrastive loss organizes graphs with similar \QAOA{} landscapes into aligned
representations, improving transfer.

\subsection{Trust-region \QAOA{} optimization}
\label{sec:trust_region}

The predicted covariance defines a constraint on the search.

\begin{definition}[Trust region]
\label{def:trust_region}
Given a predicted Gaussian $\Normal(\bm\mu,\bm\Sigma)$ for graph
$\calG$, with $\bm\Sigma$ symmetric positive definite, and a confidence
level $\alpha \in (0,1)$, the
\emph{trust region} is the Mahalanobis ellipsoid
\begin{equation}
    \calT_\alpha(\calG) = \bigl\{\bm\theta \in \bbR^{2p} :
    (\bm\theta - \bm\mu)^\top \bm\Sigma^{-1}
    (\bm\theta - \bm\mu) \le \chi^2_{2p}(\alpha)\bigr\},
    \label{eq:trust_region}
\end{equation}
where $\chi^2_{2p}(\alpha)$ is the $\alpha$-quantile of the
$\chi^2$ distribution with $2p$ degrees of freedom.
\end{definition}

Optimization is restricted to $\calT_\alpha$:
(a)~candidate angles
$\bm\theta_1,\ldots,\bm\theta_K \sim \Normal(\bm\mu,\bm\Sigma)$
are sampled from the corresponding truncated Gaussian restricted to
$\calT_\alpha$;
(b)~during Nelder--Mead refinement, any simplex vertex
$\bm\theta'$ exiting $\calT_\alpha$ is projected back:
\begin{equation}
    \mathrm{proj}_{\calT}(\bm\theta') = \bm\mu +
    \min\!\Bigl(1,\;
    \frac{\sqrt{\chi^2_{2p}(\alpha)}}
    {\|\bm\Sigma^{-1/2}(\bm\theta' - \bm\mu)\|_2}
    \Bigr)(\bm\theta' - \bm\mu).
    \label{eq:projection}
\end{equation}

This constrains the entire optimization trajectory, rather than only
its initialization~\cite{egger2021,sack2021}.

\begin{figure}[t]
    \centering
    \includegraphics[width=\columnwidth]{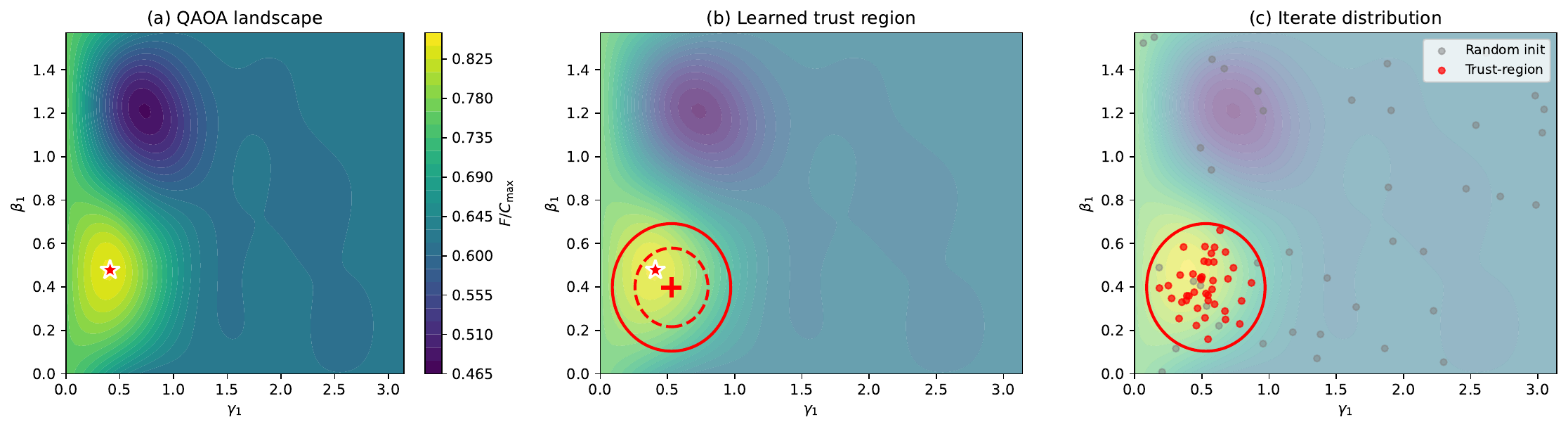}
    \caption{Trust-region mechanism on an $n{=}8$ Erd\H{o}s--R\'enyi
        instance at $p{=}2$ (layer-1 slice). (a)~Landscape of the \QAOA{}
        objective, with the optimum marked by a star. (b)~Learned
        Mahalanobis trust regions at $95\%$ (solid) and $68\%$ (dashed),
        concentrated around the high-quality basin. (c)~Random
        initializations (gray) versus trust-region samples (red), showing
        that the learned ellipsoid concentrates evaluations near the
        optimum.
    \label{fig:landscape_trust}}
\end{figure}

\Cref{fig:landscape_trust} makes the trust-region idea concrete on an
$n{=}8$ Erd\H{o}s--R\'enyi instance at $p{=}2$. The main point is that the
predicted covariance localizes the search near a productive basin of
the landscape instead of letting the optimizer spend effort across a
much broader low-value region. In that sense, the covariance acts as a
usable geometric prior rather than a passive uncertainty label.

\subsection{Uncertainty-guided adaptive refinement}
\label{sec:adaptive}

Define
\begin{equation*}
    U(\calG) = \mathrm{tr}(\bm\Sigma(\calG))/2p,
    \quad
    z(\calG) = \frac{U(\calG) - U_{\mathrm{med}}}{U_{\mathrm{iqr}}}.
\end{equation*}

Resource allocation:
\begin{align}
    K(\calG) &= \mathrm{clamp}(\lfloor 1 + 4\sigma(z)\rfloor,1,5),
    \label{eq:K_adaptive} \\
    T(\calG) &= \mathrm{clamp}(\lfloor T_{\mathrm{base}}
    (0.5 + z_+)\rfloor, 5, 2T_{\mathrm{base}}),
    \label{eq:T_adaptive}
\end{align}

Higher uncertainty increases both the number of samples and the number
of refinement steps. The mapping is monotone and bounded, enabling
comparison with fixed-budget baselines.

\Cref{fig:algorithm} reinforces the conceptual separation between
prediction and search. The learned model is used once to produce a
Gaussian, and that Gaussian then determines the local optimization
policy: where the search begins, how far it is allowed to move, and how
much computation the instance receives.

\begin{figure}[t]
\noindent\fbox{\parbox{0.95\columnwidth}{\raggedright%
\textbf{Algorithm 1: \UQQAOA{} inference}\\[4pt]%
\textbf{Input:} graph $\calG$, trained model
$f_\theta : \calG \mapsto (\bm\mu,\bm\Sigma)$,
budget $T_{\mathrm{base}}$, confidence level $\alpha$.\\[3pt]
\textbf{Output:} approximate MaxCut bitstring $z^*$.\\[3pt]
1. Predict $(\bm\mu, \bm\Sigma) \leftarrow f_\theta(\calG)$.\\
2. Compute $U \leftarrow \mathrm{tr}(\bm\Sigma)/2p$;\;
   $z \leftarrow (U - U_{\mathrm{med}})/U_{\mathrm{iqr}}$.\\
3. Set sample count
   $K \leftarrow \mathrm{clamp}(\lfloor 1 + 4\sigma(z)\rfloor,1,5)$.\\
4. Set polish budget
   $T \leftarrow \mathrm{clamp}(\lfloor T_{\mathrm{base}}(0.5+z_+)
   \rfloor, 5, 2T_{\mathrm{base}})$.\\
5. Construct trust region
   $\calT_\alpha \leftarrow \{\bm\theta :
   \|\bm\Sigma^{-1/2}(\bm\theta - \bm\mu)\|_2^2
   \le \chi^2_{2p}(\alpha)\}$.\\
6. Sample $\bm\theta_1,\ldots,\bm\theta_K$ from the truncated
    Gaussian $\Normal(\bm\mu,\bm\Sigma)$ restricted to $\calT_\alpha$.\\
7. Evaluate \QAOA{} circuit: $\bm\theta^* \leftarrow
   \arg\max_k F_\calG(\bm\theta_k)$.\\
8. Trust-region polish: run $T$ iterations of Nelder--Mead
   from $\bm\theta^*$, projecting each iterate via
   \cref{eq:projection}.\\
9. Return best bitstring $z^*$ from final \QAOA{} measurement.
}}
\caption{\UQQAOA{} inference. The predictor first converts uncertainty
into an instance-dependent sampling and refinement budget
(Steps~3--4), then performs trust-region-constrained optimization
(Steps~5--8). The total number of circuit evaluations is $K+T$, up to
constant factors from the refinement routine.
\label{fig:algorithm}}
\end{figure}

\begin{remark}[Conformal trust-region radius]
\label{rem:conformal_tr}
At inference time, the $\chi^2_{2p}(\alpha)$ radius in Step~5 can
be replaced by the conformal quantile $\hat{q}_{1-\alpha}$
(\cref{prop:conformal}), yielding a \emph{conformalized} trust
region
\begin{equation*}
    \calT_{\hat{q}_{1-\alpha}}
    = \bigl\{\bm\theta :
    (\bm\theta - \bm\mu)^\top \bm\Sigma^{-1}(\bm\theta - \bm\mu)
    \le \hat{q}_{1-\alpha}\bigr\}
\end{equation*}
with the distribution-free, finite-sample guarantee
$\bbP[\bm\theta^* \in \calT_{\hat{q}_{1-\alpha}}] \ge 1 - \alpha$.
For our validation set ($\hat{q}_{0.90} = 10.8$ vs.\
$\chi^2_4(0.90) = 7.779$), this inflates the trust region by a
factor of $(10.8/7.779)^{1/2} \approx 1.178$ in radius, i.e., an
approximately $18\%$ enlargement. This trades a larger search region
for a distribution-free finite-sample
marginal coverage statement.
\end{remark}

\section{Guarantees for Learned Trust-Region Search}
\label{sec:theory}

The following results give guarantees for the learned distribution and the induced
trust-region search. The results are local and conditional. They do not
assert global optimality of the hybrid procedure. They give
conditions under which the predicted region preserves objective
quality, supports coverage statements, and justifies uncertainty-based
resource allocation.

Throughout, $\calG \sim \calD$ denotes a held-out graph except where
otherwise stated. The predictor is
$f_\theta(\calG) = (\bm\mu(\calG),\bm\Sigma(\calG))$.
The reference point $\bm\theta^*(\calG)$ denotes the local optimum
produced by the multi-restart Nelder--Mead procedure used for target
generation, or the optimum specified in the corresponding statement.
The trust region $\calT_\alpha(\calG)$ is defined in
\cref{eq:trust_region}. When the radius is not tied to a
$\chi^2_{2p}(\alpha)$ quantile, we write $\calT_q(\calG)$.

Expectations and probabilities are taken with respect to the randomness
explicitly indicated: draws of $\calG$ from $\calD$, samples from
$\Normal(\bm\mu(\calG),\bm\Sigma(\calG))$, or uniform samples on
$\calT_\alpha(\calG)$. Constants such as $L_{\calG,K}$, $L_\calG$,
$\kappa$, $\Lambda$, and $M_3$ are deterministic once $\calG$ is fixed.

\subsection{Landscape properties}

\begin{proposition}[Local Lipschitz regularity of the \QAOA{} objective]
\label{prop:lipschitz}
For a connected graph $\calG$ with $m$ edges, the depth-$p$ \QAOA{}
objective $F_\calG : \bbR^{2p} \to \bbR$ is real analytic. In
particular, on any bounded set $K \subset \bbR^{2p}$,
\begin{equation}
    L_{\calG,K} := \sup_{\bm\theta \in K}
    \|\nabla F_\calG(\bm\theta)\|_2 < \infty.
    \label{eq:local_lipschitz_def}
\end{equation}
A conservative global bound is $L_\calG \le 2m\sqrt{2p}$.
\end{proposition}

\begin{proof}
Analyticity follows because $U_{\QAOA}(\bm\theta)$ is a finite product
of matrix exponentials. Thus $F_\calG$ is analytic and $\nabla F_\calG$
is continuous. On compact $K$, the supremum is finite, and the mean
value theorem yields the Lipschitz bound. The global estimate follows
from standard derivative bounds for \QAOA{} expectation values
\cite{cerezo2021,schuld2019}.
\end{proof}

\subsection{Trust-region guarantees}

\begin{theorem}[Lower bound on the expected objective within the trust region]
\label{thm:landscape_lb}
Let $\calT_\alpha(\calG)$ be defined by \cref{eq:trust_region}. Suppose
$F_\calG(\bm\mu) \ge r^* C_{\max}$. Let $\Theta$ be supported on
$\calT_\alpha(\calG)$. Then
\begin{equation}
    \bbE\bigl[F_\calG(\Theta)\bigr]
    \ge r^* C_{\max}
    - L_\calG \sqrt{\chi^2_{2p}(\alpha)\cdot
    \|\bm\Sigma\|_{\mathrm{op}}},
    \label{eq:landscape_lb}
\end{equation}
\end{theorem}
\begin{proof}
For $\bm\theta \in \calT_\alpha$, one has
\begin{equation*}
    \|\bm\theta - \bm\mu\|_2^2
    \le \|\bm\Sigma\|_{\mathrm{op}}\cdot
    (\bm\theta - \bm\mu)^\top \bm\Sigma^{-1}(\bm\theta - \bm\mu)
    \le \|\bm\Sigma\|_{\mathrm{op}}\cdot \chi^2_{2p}(\alpha).
\end{equation*}
Hence
\begin{equation*}
    \|\bm\theta - \bm\mu\|_2
    \le \sqrt{\chi^2_{2p}(\alpha)\|\bm\Sigma\|_{\mathrm{op}}}.
\end{equation*}
By Proposition~\ref{prop:lipschitz},
\begin{equation*}
    |F_\calG(\bm\theta) - F_\calG(\bm\mu)|
    \le L_\calG \sqrt{\chi^2_{2p}(\alpha)\|\bm\Sigma\|_{\mathrm{op}}}.
\end{equation*}
Applying the assumption $F_\calG(\bm\mu) \ge r^* C_{\max}$ and taking
expectation yields the claim.
\end{proof}

\begin{corollary}[Expected objective under variance confinement]
\label{cor:quality_confinement}
Under the conditions of Theorem~\ref{thm:landscape_lb}, if
$\sigma_{\max}^2 \le (r^* C_{\max})^2 /
(4 L_\calG^2 \chi^2_{2p}(\alpha))$, then
$\bbE[F_\calG(\Theta)] \ge \tfrac{1}{2} r^* C_{\max}$.
\end{corollary}

\begin{proof}
Set $\epsilon = \tfrac{1}{2}r^* C_{\max}$ in
Theorem~\ref{thm:landscape_lb}. This gives the sufficient condition
$\sigma_{\max}^2 \le (r^* C_{\max})^2 /
(4 L_\calG^2 \chi^2_{2p}(\alpha))$.
The conclusion follows immediately.
\end{proof}

\subsection{Gradient anti-concentration}

\begin{theorem}[Local gradient anti-concentration in the trust region]
\label{thm:anticoncentration}
Assume $F_\calG$ is three times continuously differentiable near
$\calT_\alpha(\calG)$,
$\lambda_{\min}(-\nabla^2 F_\calG(\bm\mu)) \ge \kappa > 0$,
and third derivatives are bounded by $M_3$.
Then for $\bm\theta \sim \mathrm{Unif}(\calT_\alpha)$,
\begin{equation}
    \begin{aligned}
    \mathrm{Var}_{\bm\theta \sim \mathrm{Unif}(\calT_\alpha)}
    \bigl[\partial_j F_\calG(\bm\theta)\bigr]
    &\ge \frac{\kappa^2\, \sigma_{\min}^2\, \chi^2_{2p}(\alpha)}
    {2p+2} \\
    &\quad - C_{p,\alpha}\bigl(\Lambda M_3\sigma_{\max}^3
    + M_3^2\sigma_{\max}^4\bigr),
    \end{aligned}
    \label{eq:anticoncentration}
\end{equation}
where $\Lambda := \|\nabla^2 F_\calG(\bm\mu)\|_{\mathrm{op}}$,
$\sigma_{\min} = \min_j \sigma_j$, and
$\sigma_{\max} = \max_j \sigma_j$.
\end{theorem}

\begin{proof}
See \cref{app:proof_anticonc}.
\end{proof}

\subsection{Algorithmic guarantees}

\begin{proposition}[Feasibility and monotonic best-so-far value under projection]
\label{prop:nondeg}
Projected optimization within $\calT_\alpha$ produces feasible iterates
and a nondecreasing best-so-far objective. The final iterate dominates
the initial sampled candidates. Let $\{\bm\theta_t\}_{t=0}^T$ be the
feasible iterates generated by projected Nelder--Mead within
$\calT_\alpha$, define $b_t := \max_{0 \le s \le t} F_\calG(\bm\theta_s)$,
and let $\widehat{\bm\theta}_t$ denote any feasible iterate attaining
$b_t$. Then:
\begin{enumerate}[label=(\roman*)]
\item $\bm\theta_t \in \calT_\alpha$ for all $t$.
\item $b_t$ is non-decreasing in $t$.
\item $F_\calG(\widehat{\bm\theta}_T) \ge \max_k F_\calG(\bm\theta_k)$
where $\bm\theta_1,\ldots,\bm\theta_K$ are the initial feasible samples
used to seed the search.
\end{enumerate}
\end{proposition}

\begin{proof}
(i)~The projection operator~\cref{eq:projection} maps every point
to $\calT_\alpha$: if $\bm\theta' \in \calT_\alpha$, the projection
is the identity; otherwise, it rescales
$\bm\theta' - \bm\mu$ to the boundary of $\calT_\alpha$.
(ii)~Monotonicity is immediate from the definition
$b_t = \max_{0 \le s \le t} F_\calG(\bm\theta_s)$: the index set on
which the maximum is taken only enlarges with~$t$.
(iii)~Because the initial sampled points are among the feasible points
seen by the algorithm before refinement begins,
$b_T \ge \max_{1 \le k \le K} F_\calG(\bm\theta_k)$. Since
$\widehat{\bm\theta}_T$ attains $b_T$, the claim follows.
\end{proof}

\begin{corollary}[Output guarantee under trust-region capture]
\label{cor:output_guarantee}
Let $\bm\theta_{\mathrm{out}}$ denote the best feasible angle vector returned by
Algorithm~\ref{fig:algorithm}, and let $z_{\mathrm{out}}$ be the final
bitstring sampled from the \QAOA{} circuit at
$\bm\theta_{\mathrm{out}}$. Fix a trust region $\calT_q(\calG)$ and
assume the reference local optimum $\bm\theta^*(\calG)$ belongs to that
region. Then
\begin{equation}
    F_\calG(\bm\theta_{\mathrm{out}})
    \ge F_\calG\bigl(\bm\theta^*(\calG)\bigr)
    - 2L_\calG\sqrt{q\,\|\bm\Sigma(\calG)\|_{\mathrm{op}}},
    \label{eq:output_angle_guarantee}
\end{equation}
and therefore
\begin{equation}
    \begin{aligned}
    \bbE\bigl[C(z_{\mathrm{out}})
    \mid \bm\theta^*(\calG) \in \calT_q(\calG)\bigr]
    &\ge F_\calG\bigl(\bm\theta^*(\calG)\bigr) \\
    &\quad - 2L_\calG\sqrt{q\,\|\bm\Sigma(\calG)\|_{\mathrm{op}}}.
    \end{aligned}
    \label{eq:output_bitstring_guarantee}
\end{equation}
If, in addition, the trust region has capture probability
$p_{\mathrm{cap}}(\calG)$, then since MaxCut costs are nonnegative,
\begin{equation}
    \bbE\bigl[C(z_{\mathrm{out}})\bigr]
    \ge p_{\mathrm{cap}}(\calG)
    \Bigl[F_\calG\bigl(\bm\theta^*(\calG)\bigr)
    - 2L_\calG\sqrt{q\,\|\bm\Sigma(\calG)\|_{\mathrm{op}}}\Bigr].
    \label{eq:output_unconditional}
\end{equation}
For the algorithmic $\chi^2$ region one may take
$p_{\mathrm{cap}} \ge \alpha-\eta$ from
Proposition~\ref{prop:trust_region_adequacy}; for the conformalized
region one has $p_{\mathrm{cap}} \ge 1-\alpha$.
\end{corollary}

\begin{proof}
Under the capture event, every sampled candidate and every projected
iterate lies in $\calT_q(\calG)$. Proposition~\ref{prop:trust_region_adequacy}
therefore gives the pointwise bound
\begin{equation*}
    F_\calG(\bm\theta)
    \ge F_\calG\bigl(\bm\theta^*(\calG)\bigr)
    - 2L_\calG\sqrt{q\,\|\bm\Sigma(\calG)\|_{\mathrm{op}}}
    \qquad \forall\,\bm\theta \in \calT_q(\calG).
\end{equation*}
Since Algorithm~\ref{fig:algorithm} returns an angle vector inside that
region, \cref{eq:output_angle_guarantee} follows. By definition of the
\QAOA{} objective,
$\bbE[C(z_{\mathrm{out}})\mid \bm\theta_{\mathrm{out}}]
= F_\calG(\bm\theta_{\mathrm{out}})$,
which gives \cref{eq:output_bitstring_guarantee}. The unconditional
bound \cref{eq:output_unconditional} follows by conditioning on the
capture event and using $C(z_{\mathrm{out}}) \ge 0$ for MaxCut.
\end{proof}

\subsection{Sampling guarantees}

\begin{theorem}[Best-of-$K$ guarantee]
\label{thm:bestofk}
Let $\bm\theta_1,\ldots,\bm\theta_K \stackrel{\text{i.i.d.}}{\sim}
\Normal(\bm\mu,\bm\Sigma)$. Then
\begin{equation}
    \bbE\bigl[\max_k F_\calG(\bm\theta_k)\bigr]
    \ge F_\calG(\bm\mu) - \frac{L_\calG\sqrt{\mathrm{tr}(\bm\Sigma)}}
    {\sqrt{K}}.
    \label{eq:bestofk}
\end{equation}
\end{theorem}

\begin{proof}
See \cref{app:proof_bestofk}.
\end{proof}

\begin{lemma}[Trust-region volume ratio]
\label{lem:dim_red}
For a diagonal covariance $\bm\Sigma = \mathrm{diag}(\sigma_1^2,
\ldots,\sigma_{2p}^2)$ with confidence $\alpha{=}0.95$ and $p{=}2$,
the trust-region volume relative to the unconstrained search
hypercube $[-\pi,\pi]^{2p}$ is
\begin{equation}
    \frac{\mathrm{Vol}(\calT_{0.95})}{\mathrm{Vol}([-\pi,\pi]^{2p})}
    = \frac{\pi^p [\chi^2_{4}(0.95)]^p}
    {\Gamma(p{+}1)\,(2\pi)^{2p}}
    \prod_{j=1}^{4}\sigma_j.
    \label{eq:vol_ratio}
\end{equation}
If one uses the isotropic proxy $\sigma_j \equiv \bar\sigma = 0.15$,
then \cref{eq:vol_ratio} gives
$0.2850\,\bar\sigma^4 \approx 1.44 \times 10^{-4}$.
\end{lemma}

\begin{proof}
The trust region is a 4-dimensional ellipsoid with half-lengths
$\sigma_j \sqrt{\chi^2_4(0.95)}$.
Its volume is $\frac{\pi^2}{2}[\chi^2_4(0.95)]^2
\prod_j \sigma_j$,
and the hypercube volume is $(2\pi)^4$.
The ratio follows by direct computation with $\chi^2_4(0.95)
\approx 9.49$.
\end{proof}

\subsection{Generalization and calibration}

The remaining results address out-of-sample transfer and calibration of
the learned distribution.

\begin{proposition}[Wasserstein continuity in a WL-compatible pseudometric]
\label{prop:wasserstein_lip}
If the \GIN{} encoder $\phi$ is $L_{\mathrm{enc}}$-Lipschitz
with respect to a WL-compatible graph pseudometric $d_{\mathrm{WL}}$
and the prediction heads are $L_{\mathrm{head}}$-Lipschitz, then:
\begin{equation}
    W_2\bigl(f_\theta(\calG),f_\theta(\calG')\bigr)
    \le \sqrt{2}\, L_{\mathrm{head}} L_{\mathrm{enc}}\,
    d_{\mathrm{WL}}(\calG,\calG'),
    \label{eq:wlip}
\end{equation}
where $d_{\mathrm{WL}}(\calG,\calG')=0$ is allowed for
non-isomorphic graphs that are indistinguishable by the relevant WL
statistics.
\end{proposition}

\begin{proof}
See \cref{app:proof_wlip}.
\end{proof}

\begin{theorem}[Adaptive-allocation guarantee under calibration and monotone difficulty]
\label{thm:regret}
Assume approximate calibration and the regret envelope
\cref{eq:monotone_difficulty_model}. Suppose the predicted uncertainty
satisfies
$\eta$-approximate calibration:
\begin{equation}
    \bigl|\bbP[\|\bm\theta^* - \bm\mu\|_{\bm\Sigma^{-1}} \le r]
    - F_{\chi^2_{2p}}(r^2)\bigr| \le \eta
    \quad \forall\, r > 0,
    \label{eq:eta_cal}
\end{equation}
where $F_{\chi^2_{2p}}$ is the $\chi^2_{2p}$ CDF\@.
Assume the residual optimization regret satisfies the instance-wise
envelope
\begin{equation}
    R_\pi(\calG) \le c_0 + c_1\frac{\sqrt{U(\calG)+\eta}}
    {\sqrt{K_\pi(\calG)}}
    \label{eq:monotone_difficulty_model}
\end{equation}
where $U(\calG)=\mathrm{tr}(\bm\Sigma(\calG))/2p$ and
$c_0,c_1 \ge 0$ are independent of $\pi$. Define population regret by
\begin{equation}
    \mathrm{Regret}(\pi) := \bbE_{\calG\sim\calD}[R_\pi(\calG)].
    \label{eq:policy_regret_def}
\end{equation}
Let $\bar K := \bbE_\calD[K_{\mathrm{adapt}}(\calG)]$, and let
$\pi_{\mathrm{unif}}$ denote the mean-budget-matched uniform comparator
with $K_{\mathrm{unif}}(\calG) \equiv \bar K$.
Assume further that the adaptive allocation rule satisfies the
envelope-dominance condition
\begin{equation}
    \bbE_\calD\!\left[
    \frac{\sqrt{U(\calG)+\eta}}{\sqrt{K_{\mathrm{adapt}}(\calG)}}
    \right]
    \le \frac{1}{\sqrt{\bar K}}
    \bbE_\calD\!\left[\sqrt{U(\calG)+\eta}\right].
    \label{eq:allocation_dominance}
\end{equation}
Then the adaptive allocation strategy $\pi_{\mathrm{adapt}}$
(Eqs.~\ref{eq:K_adaptive}--\ref{eq:T_adaptive}) satisfies
\begin{align}
    \mathrm{Regret}(\pi_{\mathrm{adapt}})
    &\le c_0 + c_1\,\bbE_\calD\!\left[
    \frac{\sqrt{U(\calG)+\eta}}{\sqrt{K_{\mathrm{adapt}}(\calG)}}
    \right] \nonumber \\
    &\le c_0 + \frac{c_1}{\sqrt{\bar K}}
    \bbE_\calD\!\left[\sqrt{U(\calG)+\eta}\right],
    \label{eq:regret} \\
    \bbE[B_{\pi_{\mathrm{adapt}}}]
    &= \alpha\,\bbE[B_{\pi_{\mathrm{unif}}}],
    \label{eq:budget_savings}
\end{align}
where $B_\pi$ is the total evaluation budget under policy~$\pi$ and
$\alpha = \bbE_\calD[B_{\pi_{\mathrm{adapt}}}] /
\bbE_\calD[B_{\pi_{\mathrm{unif}}}]$ depends on the deployment
distribution of predicted uncertainties. Thus $\alpha<1$ is not a
consequence of calibration alone; it depends on the deployment
distribution together with the implemented allocation rule.
\end{theorem}

\begin{proof}
See \cref{app:proof_regret}.
\end{proof}

\begin{proposition}[Monotonicity of the implemented allocation rule]
\label{prop:allocation_support}
Let
$z(\calG) = (U(\calG)-U_{\mathrm{med}})/U_{\mathrm{iqr}}$, and define
$K_{\mathrm{adapt}}$ and $T_{\mathrm{adapt}}$ by
\cref{eq:K_adaptive,eq:T_adaptive}. Then both allocations are
nondecreasing functions of $U$, and one has the explicit thresholds
\begin{equation}
    K_{\mathrm{adapt}}(\calG) =
    \begin{cases}
        1, & z(\calG) < -\log 3, \\
        2, & -\log 3 \le z(\calG) < 0, \\
        3, & 0 \le z(\calG) < \log 3, \\
        4, & z(\calG) \ge \log 3.
    \end{cases}
    \label{eq:adaptive_thresholds}
\end{equation}
If $T_{\mathrm{base}} \ge 10$, then
$T_{\mathrm{adapt}}(\calG)=\lfloor T_{\mathrm{base}}/2\rfloor$ for
$z(\calG) \le 0$, and $T_{\mathrm{adapt}}$ increases linearly in
$z(\calG)$ until it reaches the cap $2T_{\mathrm{base}}$.
Thus the implemented policy reallocates computation monotonically from
the low-$U$ region to the high-$U$ region, which is the ordering premise
behind \cref{eq:allocation_dominance}. The same ordering is supported
empirically by \cref{fig:reliability,tab:efficiency}: larger predicted
uncertainty is associated with harder instances, and exploiting that
ordering lowers the total query count.
\end{proposition}

\begin{proof}
The sigmoid is strictly increasing, so
$\lfloor 1+4\sigma(z)\rfloor$ is nondecreasing in $z$.
The thresholds follow from the identities
$\sigma(z)=1/4$ at $z=-\log 3$,
$\sigma(z)=1/2$ at $z=0$, and
$\sigma(z)=3/4$ at $z=\log 3$.
Because $\sigma(z) < 1$ for finite $z$, the implemented rule takes the
values $1,2,3,4$ on the four regimes in
\cref{eq:adaptive_thresholds}. The monotonicity of
$T_{\mathrm{adapt}}$ follows in the same way from the monotonicity of
$z_+ = \max(z,0)$ and the clamp map. If $z \le 0$, then
$z_+=0$ and
$T_{\mathrm{adapt}} = \mathrm{clamp}(\lfloor T_{\mathrm{base}}/2\rfloor,
5,2T_{\mathrm{base}}) = \lfloor T_{\mathrm{base}}/2\rfloor$
whenever $T_{\mathrm{base}} \ge 10$.
The empirical statements are exactly those reported in
\cref{fig:reliability,tab:efficiency}.
\end{proof}

\begin{theorem}[Spectral-norm generalization guarantee]
\label{thm:generalization}
Let $f_\theta$ be the trained \GIN{} predictor with $L$ layers,
hidden dimension~$d$, and per-layer spectral norms
$s_\ell = \|W_\ell\|_{\sigma}$.
Let $B_x = \max_{\calG} \|\bm{h}^{(0)}\|_2$ be the bound on
input features. Assume the induced loss class
$\calF = \{W_2^2(f_\theta(\cdot),\delta_{\bm\theta^*(\cdot)})\}$ is
uniformly bounded by $M$ and satisfies the empirical Rademacher
complexity estimate
\begin{equation*}
    \mathfrak{R}_N(\calF) \le \frac{2 B_x \prod_{\ell=1}^{L} s_\ell
    \sqrt{2L\log(2d)}}{\sqrt{N}}.
\end{equation*}
Given $N$ i.i.d.\ training graphs, the
population Wasserstein risk
$\calR(f_\theta)
\!:=\!
\bbE_{\calG \sim \calD}[W_2^2(f_\theta(\calG),
\delta_{\bm\theta^*(\calG)})]$
satisfies, with probability $\ge 1 - \delta$ over the
training set:
\begin{align}
    \calR(f_\theta)
    &\le \hat{\calR}_N
    + \frac{4 B_x \prod_{\ell=1}^{L} s_\ell}{\sqrt{N}}
    \sqrt{2L\log(2d)} \nonumber \\
    &\quad + 3M\!\sqrt{\frac{\log(2/\delta)}{2N}},
    \label{eq:gen_bound}
\end{align}
where $\hat{\calR}_N$ is the empirical Wasserstein risk
and $M$ is the loss bound. Here $\delta_{\bm\theta^*(\calG)}$ denotes
the Dirac point mass at the target angle vector. The complexity term scales with
$\prod_\ell s_\ell \sqrt{L\log d / N}$ rather than with the total
parameter count~\cite{bartlett2017,neyshabur2018,golowich2018}.
\end{theorem}

\begin{proof}
See \cref{app:proof_gen}.
\end{proof}

\begin{corollary}[Sample-complexity consequence]
\label{cor:sample_complexity}
Under the conditions of Theorem~\ref{thm:generalization}, to
achieve population risk within $\epsilon$ of the empirical risk
with probability $\ge 1 - \delta$, it suffices to train on
\begin{equation}
    N \ge O\!\left(
    \frac{B_x^2 \bigl(\prod_\ell s_\ell\bigr)^2
    L \log d}{\epsilon^2}
    + \frac{M^2 \log(1/\delta)}{\epsilon^2}
    \right)
    \label{eq:sample_complexity}
\end{equation}
graphs.
\end{corollary}

\begin{proof}
Set the generalization gap
\begin{equation*}
    \frac{4 B_x \prod s_\ell \sqrt{2L\log(2d)}}{\sqrt{N}}
    + 3M\sqrt{\frac{\log(2/\delta)}{2N}}
    \le \epsilon
\end{equation*}
and solve for $N$.
A sufficient condition is that both terms are bounded by $\epsilon/2$, which holds when
$N \ge \frac{128 B_x^2 (\prod s_\ell)^2 L \log(2d)}{\epsilon^2}$
and $N \ge \frac{18 M^2 \log(2/\delta)}{\epsilon^2}$
respectively.
\end{proof}

\subsection{Conformal coverage}

\begin{proposition}[Finite-sample conformal coverage]
\label{prop:conformal}
Let $\{(\calG_i, \bm\theta^*_i)\}_{i=1}^M$ be an exchangeable
validation set.  Define the nonconformity score
\begin{equation*}
    s(\calG)
    = (\bm\theta^*(\calG) - \bm\mu(\calG))^\top
    \bm\Sigma^{-1}(\calG)\,(\bm\theta^*(\calG) - \bm\mu(\calG)).
\end{equation*}
Let $k = \lceil(M+1)(1-\alpha)\rceil$, and let
$\hat{q}_{1-\alpha}$ be the $k$-th order statistic of
$\{s(\calG_i)\}_{i=1}^M$. Then for any new exchangeable test
instance $\calG_{M+1}$:
\begin{equation}
    \bbP\bigl[\bm\theta^*(\calG_{M+1}) \in
    \calT_{\hat{q}_{1-\alpha}}(\calG_{M+1})\bigr] \ge 1 - \alpha,
    \label{eq:conformal_coverage}
\end{equation}
where
\begin{equation*}
    \calT_q(\calG)
    = \bigl\{\bm\theta :
    (\bm\theta - \bm\mu)^\top \bm\Sigma^{-1}(\bm\theta - \bm\mu)
    \le q\bigr\}
\end{equation*}
~\cite{vovk2005,romano2019,angelopoulos2021}.
The coverage statement applies to the target optima $\bm\theta_i^*$
used throughout the paper.
\end{proposition}

\begin{proof}
By the exchangeability of $\{(\calG_i,\bm\theta^*_i)\}_{i=1}^{M+1}$,
the rank of $s(\calG_{M+1})$ among
$\{s(\calG_1),\ldots,s(\calG_{M+1})\}$ is uniformly distributed
on $\{1,\ldots,M+1\}$.  Therefore
$\bbP[s(\calG_{M+1}) \le \hat{q}_{1-\alpha}]
= \bbP[\mathrm{rank}(s(\calG_{M+1})) \le k]
\ge k/(M+1) \ge 1-\alpha$.
Since $s(\calG) \le q \Leftrightarrow
\bm\theta^* \in \calT_q(\calG)$,
the result follows~\cite{vovk2005,angelopoulos2021}.
\end{proof}

\begin{proposition}[Trust-region capture and local objective control under calibration]
\label{prop:trust_region_adequacy}
Let $\bm\theta^*(\calG)$ denote the reference local optimum for graph
$\calG$, and let $\Theta$ be any random variable supported on a trust
region centered at the learned mean $\bm\mu(\calG)$.

\emph{(a) Algorithmic $\chi^2$ region.}
If the Mahalanobis score satisfies the $\eta$-approximate calibration
condition in \cref{eq:eta_cal}, then for the trust region
$\calT_\alpha(\calG)$ from \cref{def:trust_region},
\begin{equation}
    \bbP\bigl[\bm\theta^*(\calG) \in \calT_\alpha(\calG)\bigr]
    \ge \alpha - \eta.
    \label{eq:trust_capture_chi2}
\end{equation}

\emph{(b) Conformal region.}
For the conformalized trust region $\calT_{\hat q_{1-\alpha}}(\calG)$ from
Proposition~\ref{prop:conformal},
\begin{equation}
    \bbP\bigl[\bm\theta^*(\calG) \in \calT_{\hat q_{1-\alpha}}(\calG)\bigr]
    \ge 1-\alpha.
    \label{eq:trust_capture_conformal}
\end{equation}

\emph{(c) Local objective control after capture.}
Let
\begin{equation}
    R(\calG;q) := \sqrt{q\,\|\bm\Sigma(\calG)\|_{\mathrm{op}}}.
    \label{eq:trust_radius_generic}
\end{equation}
On the event $\bm\theta^*(\calG) \in \calT_q(\calG)$, every
$\bm\theta \in \calT_q(\calG)$ satisfies
\begin{equation}
    \|\bm\theta - \bm\theta^*(\calG)\|_2 \le 2R(\calG;q),
    \label{eq:trust_two_radius}
\end{equation}
and therefore, by Lipschitz continuity,
\begin{equation}
    F_\calG(\bm\theta)
    \ge F_\calG\bigl(\bm\theta^*(\calG)\bigr)
    - 2L_\calG R(\calG;q).
    \label{eq:trust_target_gap}
\end{equation}
Consequently,
\begin{equation}
    \bbE\Bigl[F_\calG(\Theta)
    \mid \bm\theta^*(\calG) \in \calT_q(\calG)\Bigr]
    \ge F_\calG\bigl(\bm\theta^*(\calG)\bigr)
    - 2L_\calG R(\calG;q).
    \label{eq:trust_target_gap_exp}
\end{equation}
\end{proposition}

\begin{proof}
For part~(a), apply \cref{eq:eta_cal} with
$r = \sqrt{\chi^2_{2p}(\alpha)}$. Since
$F_{\chi^2_{2p}}(\chi^2_{2p}(\alpha)) = \alpha$,
\begin{align*}
    \bbP\bigl[\bm\theta^*(\calG) \in \calT_\alpha(\calG)\bigr]
    &= \bbP\Bigl[\|\bm\theta^*(\calG)-\bm\mu(\calG)\|_{\bm\Sigma^{-1}}^2
    \le \chi^2_{2p}(\alpha)\Bigr] \\
    &\ge \alpha - \eta.
\end{align*}
Part~(b) is exactly Proposition~\ref{prop:conformal}.

For part~(c), if both $\bm\theta$ and $\bm\theta^*(\calG)$ belong to
$\calT_q(\calG)$, then by the Euclidean radius bound associated with
\cref{eq:trust_region},
\begin{equation*}
    \|\bm\theta-\bm\mu\|_2 \le R(\calG;q),
    \qquad
    \|\bm\theta^*(\calG)-\bm\mu\|_2 \le R(\calG;q).
\end{equation*}
The triangle inequality gives \cref{eq:trust_two_radius}. Applying
Proposition~\ref{prop:lipschitz} around the reference local optimum then
yields
\begin{equation*}
    \begin{aligned}
    F_\calG(\bm\theta)
    &\ge F_\calG\bigl(\bm\theta^*(\calG)\bigr)
    - L_\calG\|\bm\theta-\bm\theta^*(\calG)\|_2 \\
    &\ge F_\calG\bigl(\bm\theta^*(\calG)\bigr)
    - 2L_\calG R(\calG;q),
    \end{aligned}
\end{equation*}
which is \cref{eq:trust_target_gap}. Taking conditional expectation over
any $\Theta$ supported on $\calT_q(\calG)$ gives
\cref{eq:trust_target_gap_exp}.
\end{proof}

\begin{remark}[Finite-scale sharpness]
\label{rem:bound_tightness}
The worst-case Lipschitz constant
$L_\calG = 2m\sqrt{2p}$ (Proposition~\ref{prop:lipschitz})
grows linearly with edges.
At $n{=}14$ for ER graphs ($m \approx 46$),
the crude global prefactor in Theorem~\ref{thm:landscape_lb} is
$L_\calG = 184$, so the lower bound remains vacuous at this scale
unless $\sqrt{\chi^2_4(0.95)\,\|\bm\Sigma\|_{\mathrm{op}}}$ is well
below $1$.
The spectral-norm generalization bound
(Theorem~\ref{thm:generalization}) is loose
($\hat{\calR}_N + 7.1$) but tighter than the na\"ive
covering-number estimate, with the correct scaling in
spectral norms and layer count rather than parameter count.
For bounded-degree graphs ($d$-regular, $m = nd/2$), the
landscape bound scales as $r^* - O(d\sqrt{p/n})$ when
combined with $\sigma_{\max} = O(1/\sqrt{n})$~\cite{brandao2018},
becoming non-vacuous for large $n$.
Notably, the conformal guarantee
(Proposition~\ref{prop:conformal}) provides a
\textbf{non-vacuous, distribution-free} coverage bound at
the experimental scale:
$\bbP[\bm\theta^* \in \calT_{\hat{q}_{0.90}}] \ge 0.90$
for the actual validation data.
Tighter landscape bounds leveraging local
curvature~\cite{larocca2023,mcclean2018} are a priority.
\end{remark}

\subsection{Noise robustness}
\label{sec:noise_theory}

We analyze the learned trust region under depolarizing noise, the
benchmark noise model used throughout the paper~\cite{wang2021noise,preskill2018}.

\begin{theorem}[Preservation of expected ordering under depolarizing noise]
\label{thm:noise_robust}
Consider a global depolarizing noise model where the noisy \QAOA{}
state after depth~$p$ satisfies
\begin{equation}
    \rho^{\varepsilon} = (1 - \nu)\,
    \ket{\bm\theta}\!\bra{\bm\theta}
    + \nu\,\frac{I}{2^n},
    \label{eq:noisy_state}
\end{equation}
with noise strength $\nu = 1 - (1-\varepsilon)^{2p}$,
$\varepsilon \in [0,1]$.
The noisy objective is
\begin{equation}
    F_\calG^\varepsilon(\bm\theta)
    = (1 - \nu)\,F_\calG(\bm\theta) + \nu\,\bar{C},
    \label{eq:noisy_obj}
\end{equation}
where $\bar{C} = \mathrm{tr}(C)/2^n = m/2$ is the
expected MaxCut cost at the maximally mixed state.
Let $\Theta_{\mathrm{TR}}$ be any random variable supported on
$\calT_\alpha(\calG)$ almost surely, and let
$\Theta_{\mathrm{rand}}$ denote a random initialization baseline.
Assume also that the predicted mean satisfies
$F_\calG(\bm\mu) \ge r^* C_{\max}$ for some $r^*>0$.
Then:

\emph{(a) Expected noisy landscape bound.}
\begin{equation}
    \begin{aligned}
    \bbE\bigl[F_\calG^\varepsilon(\Theta_{\mathrm{TR}})\bigr]
    &\ge (1-\nu)\bigl[r^* C_{\max}
    - L_\calG \sqrt{\chi^2_{2p}(\alpha)\,\|\bm\Sigma\|_{\mathrm{op}}}\bigr] \\
    &\quad + \nu\,\bar{C}.
    \end{aligned}
    \label{eq:noisy_lb}
\end{equation}

\emph{(b) Preserved expected ordering.}
\begin{equation}
    \begin{aligned}
    \bbE\bigl[F_\calG^\varepsilon(\Theta_{\mathrm{TR}})\bigr]
    - \bbE\bigl[F_\calG^\varepsilon(\Theta_{\mathrm{rand}})\bigr]
    &= (1-\nu)\Bigl(\bbE\bigl[F_\calG(\Theta_{\mathrm{TR}})\bigr] \\
    &\qquad - \bbE\bigl[F_\calG(\Theta_{\mathrm{rand}})\bigr]\Bigr).
    \end{aligned}
    \label{eq:noise_ordering}
\end{equation}
Hence any noiseless expected advantage is preserved for $\varepsilon<1$.

\emph{(c) Preserved relative expected advantage.}
\begin{equation}
    \frac{\bbE[F_\calG^\varepsilon(\Theta_{\mathrm{TR}})]
    - \bar{C}}{\bbE[F_\calG^\varepsilon(\Theta_{\mathrm{rand}})]
    - \bar{C}}
    = \frac{\bbE[F_\calG(\Theta_{\mathrm{TR}})]
    - \bar{C}}{\bbE[F_\calG(\Theta_{\mathrm{rand}})]
    - \bar{C}},
    \label{eq:noise_amplify}
\end{equation}
The ratio statement in part~(c) is understood whenever the denominator
$\bbE[F_\calG^\varepsilon(\Theta_{\mathrm{rand}})]-\bar C$ is nonzero.
\end{theorem}

\begin{proof}
See \cref{app:proof_noise}.
\end{proof}

\section{Experimental Protocol}
\label{sec:setup}

The experiments test how the learned trust region affects query efficiency and
solution quality. The graph families are chosen to
span distinct structural regimes, and the baselines separate heuristic
transfer, learned initialization, and distributional prediction.
Evaluation count is treated as the primary metric, while sampled
approximation ratio serves as a constraint on solution quality.

\textbf{Problem instances.}
Unweighted MaxCut at depth $p{=}2$ is evaluated on four graph families:
Erd\H{o}s--R\'enyi ER($n,0.5$)~\cite{erdos1959}, 3-regular,
Barab\'asi--Albert BA($n,2$)~\cite{barabasi1999}, and
Watts--Strogatz WS($n,4,0.3$)~\cite{watts1998}.
The test set uses $n \in \{8,10,12,14,16\}$ with 48 instances per size
(12 per family). The families differ in degree distribution,
connectivity, and clustering, providing a test of transfer beyond the
training distribution.

\textbf{Training.}
Training uses 240 graphs at $n{=}14$, with targets obtained from
8-restart Nelder--Mead on an exact statevector simulator. The validation
set contains 80 graphs. Training at a single graph size separates
cross-family transfer from interpolation across $n$. Total training time
is approximately two minutes on a single CPU core.

\textbf{Baselines.}
We compare five methods:
\begin{enumerate}[leftmargin=*,itemsep=1pt]
\item \textbf{Random initialization:} 4 independent random restarts.
\item \textbf{Concentration heuristic}~\cite{brandao2018,wurtz2021}:
2 restarts from median training angles.
\item \textbf{$k$-NN regressor:} $k{=}5$ nearest neighbors using
8-dimensional handcrafted features
$(\rho, \bar{d}, \sigma_d, \bar{C}, \lambda_2, \lambda_n, n, m)$.
\item \textbf{TQA}~\cite{sack2021}: Trotterized quantum annealing
initialization.
\item \textbf{\GNN{} point predictor:} deterministic \GIN{} with a
single mean head.
\end{enumerate}
All methods use the same Nelder--Mead refinement with identical
tolerances (iterate tolerance $x_{\mathrm{atol}}{=}10^{-4}$ and
objective tolerance $f_{\mathrm{atol}}{=}10^{-6}$). The \GNN{} point predictor isolates the
effect of improved initialization from the use of a learned covariance
to constrain the search.

\textbf{Metrics.}
The primary metric is the number of objective evaluations.
Solution quality is measured by the sampled best-bitstring
approximation ratio
$r := C(z_{\mathrm{best}})/C_{\max}$.
Additional metrics include median wall-clock time (ms), speedup over
random initialization, evaluation reduction, efficiency-adjusted
quality, and Spearman $\rho$ for calibration.
Efficiency-adjusted quality is reported as $r/100\text{evals}$ and
$r/\text{s}$.
Unless stated otherwise, reported ratios refer to the sampled
best-bitstring ratio rather than the expected objective
$F_\calG(\bm\theta)/C_{\max}$.
Expected objectives are used separately when comparing with the
landscape guarantees in \cref{sec:theory}.
All results are reported as mean $\pm$ standard deviation across
instances. Statistical significance is assessed using Wilcoxon
signed-rank tests (\cref{app:significance}), with $p < 0.05$ for all
main comparisons.

\textbf{Quantum circuit resources.}
Each depth-$p{=}2$ MaxCut \QAOA{} circuit on $n$ qubits with $m$
edges contains $2pm = 4m$ $\mathrm{ZZ}(\gamma)$ entangling layers and
$2pn = 4n$ single-qubit $R_x(\beta)$ rotations, corresponding to
$8m + 4n$ CNOT gates and $4m + 4n$ rotation gates per evaluation.
For $n{=}14$, $m \in [21,46]$, yielding $172$--$372$ CNOT gates per
evaluation. \UQQAOA{} requires 45 evaluations, corresponding to
$7{,}740$--$16{,}740$ CNOT executions, compared to
$59{,}000$--$128{,}000$ for random initialization at 343 evaluations,
an $87\%$ reduction in total gate count.
Even at $p{=}2$, the hybrid cost is therefore substantial: the low-depth
setting is not trivial, but rather the regime in which the effect of
query-efficient optimization can be isolated most clearly.

\section{Empirical Evaluation}
\label{sec:results}

The results evaluate whether the learned Gaussian provides more than an improved
initialization for the hybrid loop. The reported quantities include
evaluation count, runtime, efficiency-adjusted quality, generalization
across graph size and family, sensitivity to training seed, and
robustness to finite-shot noise. The central questions are whether the
reduction in evaluations is attributable to the learned trust region
and allocation rule, whether this behavior persists across
distributions, and whether the predicted uncertainty is consistent with
the calibration and allocation assumptions of \cref{sec:theory}.

\subsection{Computational efficiency}
\label{sec:efficiency}

\begin{figure}[t]
    \centering
    \includegraphics[width=\columnwidth]{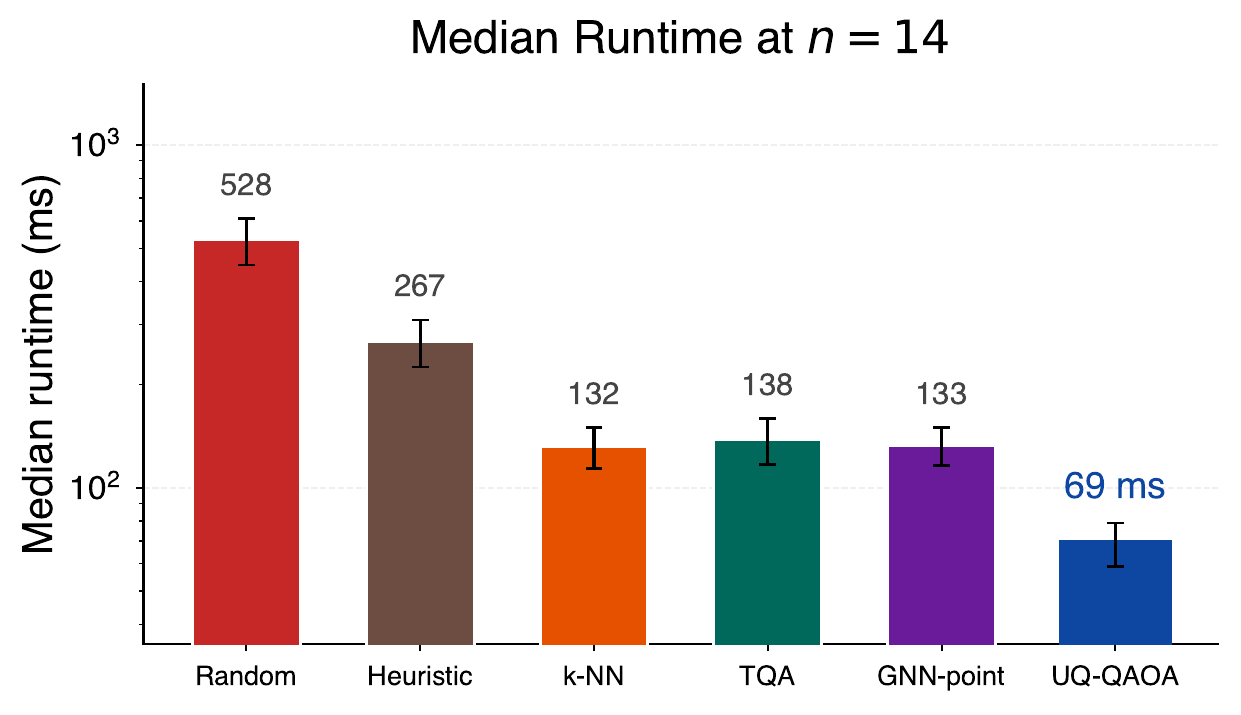}
    \caption{Median wall-clock runtime at $n{=}14$ on a logarithmic
    scale.
    \label{fig:runtime}}
\end{figure}

\begin{figure}[t]
    \centering
    \includegraphics[width=\columnwidth]{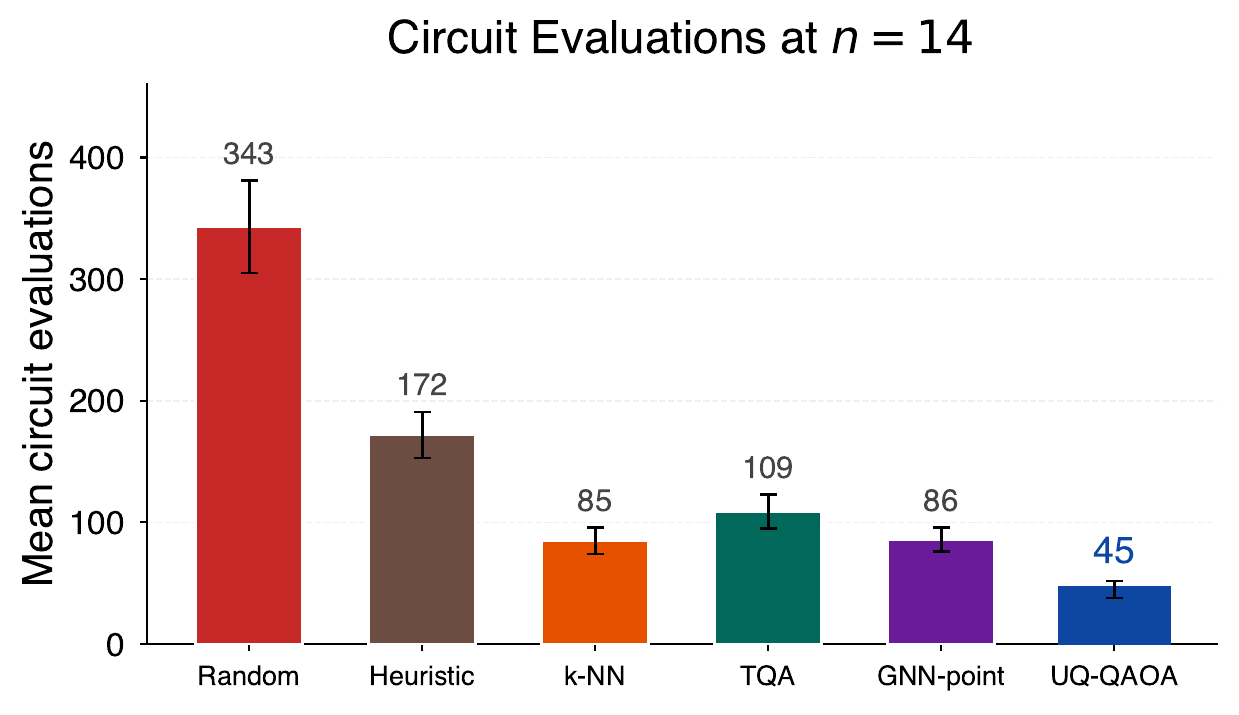}
    \caption{Mean number of circuit evaluations at $n{=}14$.
    \label{fig:evals}}
\end{figure}

\Cref{tab:efficiency} summarizes the primary tradeoff. \UQQAOA{} attains
the lowest evaluation count ($45\pm 7$), the lowest
wall-clock time ($69\pm 10$~ms), and the largest speedup
($7.7\times$). It does not achieve the highest sampled best-bitstring
ratio; the concentration heuristic and TQA remain superior in absolute
ratio. The result is therefore a reduction in query cost at comparable
solution quality. Relative to the strongest learned point baseline
($85$ evaluations), the reduction to $45$ evaluations indicates that
the covariance and adaptive allocation contribute beyond
initialization. The reduction in evaluations is not explained by
improved initialization alone. The \GNN{} point baseline yields
comparable initial-parameter quality but requires nearly twice as many
evaluations. The additional reduction arises from constraining the
search region and adapting the evaluation budget using the predicted
covariance.
What stands out in the table is that the advantage of \UQQAOA{} is
mainly computational. The strongest baselines remain slightly better in
sampled ratio, but that difference is small compared with the reduction
in evaluations and runtime. The method therefore changes the
quality-cost balance rather than merely nudging absolute solution
quality.
This same point is visible in \cref{fig:runtime,fig:evals}.
\Cref{fig:runtime} shows the wall-clock consequence on a logarithmic
scale: \UQQAOA{} sits clearly below the baselines. \Cref{fig:evals}
shows the same ordering in the quantity that matters most for hardware
use, namely the number of circuit evaluations.

\begin{table}[t]
\caption{Computational efficiency at $n{=}14$ and $p{=}2$
(48 test instances, mean$\pm$std across instances). ``Ratio'' denotes
the sampled best-bitstring approximation ratio
$C(z_{\mathrm{best}})/C_{\max}$. \textbf{Bold} marks the best value in
each column.
\label{tab:efficiency}}
\begin{ruledtabular}
\scriptsize
\begin{tabular*}{\columnwidth}{@{\extracolsep{\fill}}lcccc}
    Method & Best-bitstring ratio & Evals
    & Med.\ ms & Speedup \\
    \hline
    Random ($R{=}4$)
      & $0.831\pm .048$ & $343\pm 38$
      & $528\pm 62$ & $1.0\times$ \\
    Heuristic~\cite{brandao2018}
      & $\mathbf{0.865}\pm .024$
      & $172\pm 19$ & $267\pm 31$
      & $2.0\times$ \\
    $k$-NN
      & $0.834\pm .027$ & $85\pm 11$
      & $132\pm 17$ & $4.0\times$ \\
    TQA~\cite{sack2021}
      & $\mathbf{0.865}\pm .023$
      & $109\pm 14$ & $138\pm 18$
      & $3.8\times$ \\
    \GNN{} point
      & $0.852\pm .025$ & $86\pm 10$
      & $133\pm 16$ & $4.0\times$ \\
    \UQQAOA{} (ours)
      & $0.838\pm .022$
      & $\mathbf{45}\pm 7$
      & $\mathbf{69}\pm 10$
      & $\mathbf{7.7\times}$ \\
\end{tabular*}
\end{ruledtabular}
\end{table}

\textbf{GNN overhead.}
The \GNN{} forward pass requires approximately $1.2$\,ms ($1.7\%$ of
runtime), so total cost is dominated by circuit evaluations.

\subsection{Efficiency-adjusted quality}
\label{sec:eff_quality}

\begin{figure}[t]
    \centering
    \includegraphics[width=\columnwidth]{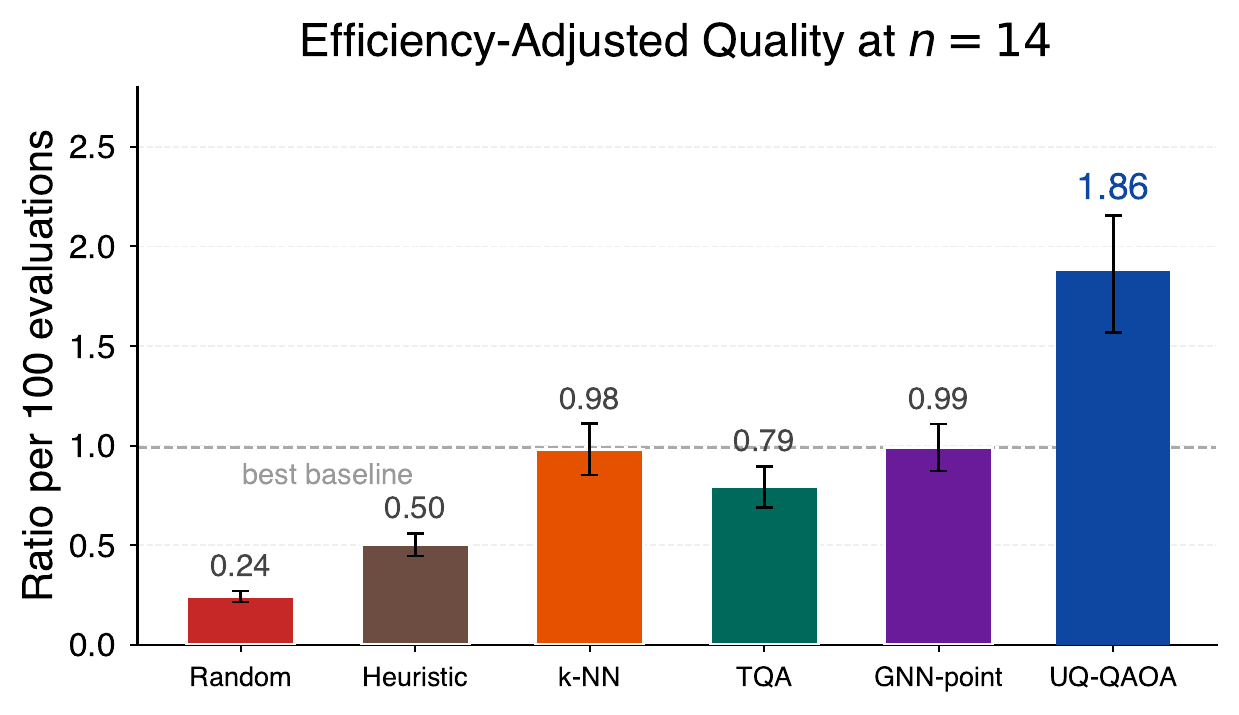}
    \caption{Efficiency-adjusted quality at $n{=}14$, measured as the
    sampled best-bitstring approximation ratio per 100 circuit
    evaluations. The dashed reference line marks the strongest
    non-\UQQAOA{} baseline, so values above one indicate that the
    method delivers better quality per fixed query budget than any
    competing baseline.
    \label{fig:eff_quality}}
\end{figure}

To account for the reduced query budget, we report quality normalized
by cost. \Cref{tab:eff_quality} shows that \UQQAOA{} attains the highest
efficiency-adjusted metrics. This table should be read together with
\cref{tab:efficiency}: the reduction in evaluations is meaningful
because solution quality remains competitive, even though the best
absolute ratios are obtained by the heuristic and TQA baselines.
Once cost is put in the denominator, the ranking changes. Methods that
look strongest in absolute ratio no longer look strongest per unit of
hardware effort. In the query-limited regime considered here, that is
the more relevant comparison, and \UQQAOA{} comes out clearly ahead.
\Cref{fig:eff_quality} shows the same shift in a more compact way:
after accounting for budget, \UQQAOA{} separates from the best
competing baseline by a visible margin.

\begin{table}[t]
\caption{Efficiency-adjusted quality at $n{=}14$ (mean$\pm$std).
These metrics summarize the tradeoff between sampled best-bitstring
quality and circuit budget rather than absolute solution quality alone.
Here $r$ denotes the sampled best-bitstring approximation ratio, and
\textbf{bold} marks the best value in each column.
\label{tab:eff_quality}}
\begin{ruledtabular}
\begin{tabular}{lccc}
    Method & $r/100\text{evals}$
    & $r/\text{s}$ & Eval.\ red. \\
    \hline
    Random ($R{=}4$)
      & $0.242\pm .015$ & $1.57\pm .18$
      & $0.0\pm 0.0\%$ \\
    Heuristic
      & $0.503\pm .031$ & $3.24\pm .38$
      & $49.9\pm 5.6\%$ \\
    $k$-NN
      & $0.981\pm .065$ & $6.32\pm .74$
      & $75.2\pm 3.1\%$ \\
    TQA
      & $0.794\pm .048$ & $6.27\pm .73$
      & $68.2\pm 4.0\%$ \\
    \GNN{} point
      & $0.991\pm .061$ & $6.41\pm .75$
      & $74.9\pm 3.0\%$ \\
    \UQQAOA{} (ours)
      & $\mathbf{1.862}\pm .098$
      & $\mathbf{12.14}\pm 1.42$
      & $\mathbf{86.9}\pm 2.1\%$ \\
\end{tabular}
\end{ruledtabular}
\end{table}

\subsection{Circuit evaluation savings}
\label{sec:eval_savings}

\Cref{fig:eval_savings} isolates circuit evaluations, which determine
hardware cost. The main impression from the figure is that \UQQAOA{}
does not merely inherit the savings of learned point prediction; it
pushes substantially further. Relative to random initialization, the
remaining query load is cut much more sharply than in the $k$-NN and
\GNN{} point baselines.

\begin{figure}[t]
    \centering
    \includegraphics[width=\columnwidth]{figures/fig5_eval_reduction.pdf}
    \caption{Reduction in circuit evaluations relative to random
    initialization at $n{=}14$. The random-initialization baseline is
    normalized to $0\%$, so each bar reports the fraction of hardware
    queries avoided by replacing uninformed initialization with a
    learned initialization strategy.
    \label{fig:eval_savings}}
\end{figure}

\subsection{Cross-size generalization}
\label{sec:crosssize}

\begin{figure}[t]
    \centering
    \includegraphics[width=\columnwidth]{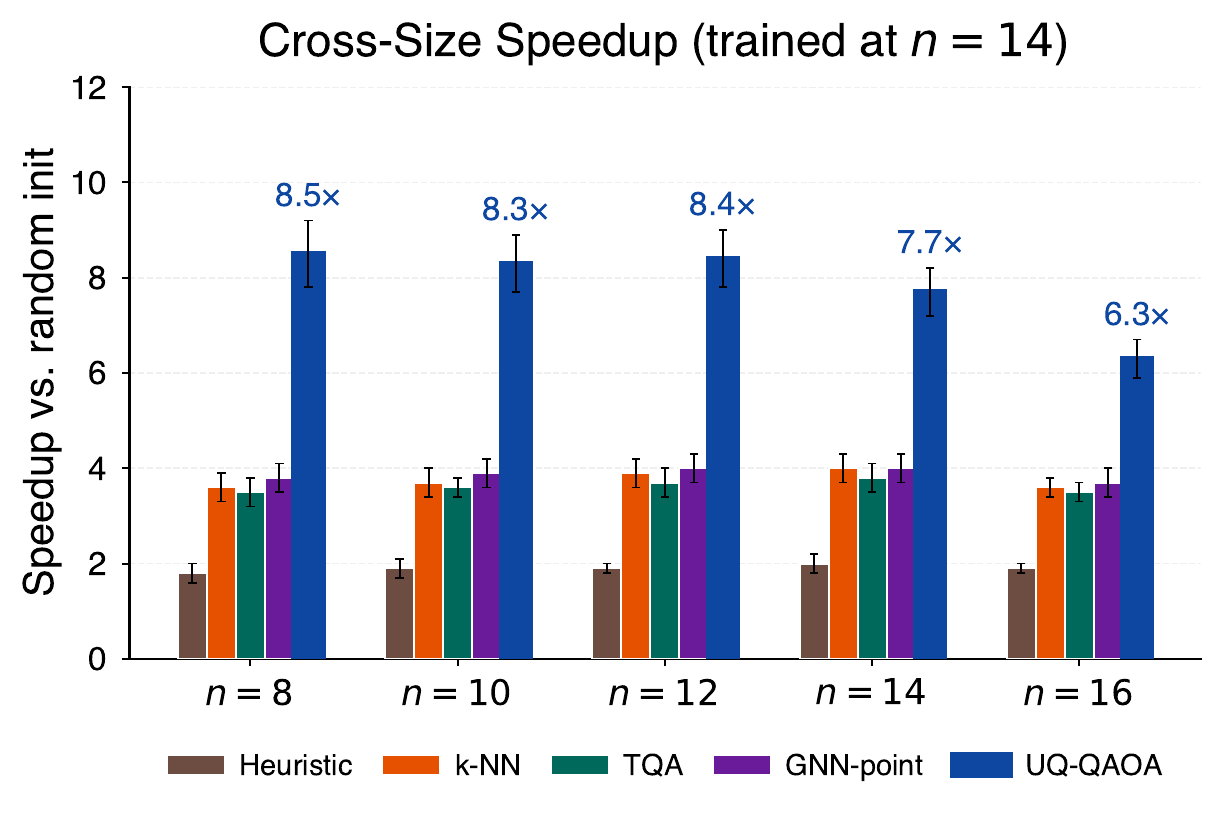}
    \caption{Speedup over random initialization across graph sizes,
    showing transfer beyond the training size.
    \label{fig:speedup}}
\end{figure}

The model is trained at $n{=}14$, as all other sizes are
out-of-distribution. \Cref{tab:crosssize} shows that the speedup
remains between $6\times$ and $9\times$ across all tested sizes,
decreasing from $8.5\times$ at $n{=}8$ to $6.3\times$ at $n{=}16$.
This indicates that the learned trust-region geometry transfers across
problem sizes without retraining.
An important feature of the table is how little the ordering changes
with graph size. The gain is not narrowly tied to the training size.
There is some weakening at $n{=}16$, which is what one should expect out
of distribution, but the basic advantage survives the change in scale.
\Cref{fig:speedup} shows the same result as a curve over graph size.
The important feature is that the curve stays high across the full
range and bends only mildly as $n$ increases. Transfer weakens at larger
size, but it does not break down.

\begin{table}[t]
\caption{Cross-size generalization, reported as speedup over random
initialization. Training is performed only at $n{=}14$, so all other
sizes are out of distribution. The table isolates whether the learned
trust-region geometry remains useful away from the training size.
Absolute \UQQAOA{} runtimes are 3/6/18/69/375\,ms for
$n{=}8$/$10$/$12$/$14$/$16$.
\label{tab:crosssize}}
\begin{ruledtabular}
\begin{tabular}{lccccc}
    & \multicolumn{5}{c}{Speedup over random initialization} \\
    \cmidrule(lr){2-6}
    Method & $n{=}8$ & $n{=}10$ & $n{=}12$ & $n{=}14$ & $n{=}16$ \\
    \hline
    Heuristic & $1.8\pm .2$ & $1.9\pm .2$ & $1.9\pm .3$ & $2.0\pm .3$ & $1.9\pm .3$ \\
    $k$-NN    & $3.6\pm .4$ & $3.7\pm .4$ & $3.9\pm .5$ & $4.0\pm .5$ & $3.6\pm .5$ \\
    TQA       & $3.5\pm .4$ & $3.6\pm .4$ & $3.7\pm .5$ & $3.8\pm .5$ & $3.5\pm .5$ \\
    \GNN{} pt & $3.8\pm .4$ & $3.9\pm .5$ & $4.0\pm .5$ & $4.0\pm .5$ & $3.7\pm .5$ \\
    \UQQAOA{}
      & $\mathbf{8.5}\pm .9$ & $\mathbf{8.3}\pm .8$ & $\mathbf{8.4}\pm .9$
      & $\mathbf{7.7}\pm .8$ & $\mathbf{6.3}\pm .7$ \\
\end{tabular}
\end{ruledtabular}
\end{table}

\subsection{Per-family analysis}
\label{sec:perfamily}

\Cref{tab:perfamily} shows that the reduction in evaluations is not
specific to a single graph family. \UQQAOA{} achieves the lowest
evaluation count and the largest speedup in each family, while the
heuristic or \GNN{} point baseline achieves the highest sampled
best-bitstring ratio.
What is striking in the table is how consistently the same pattern
reappears. In every family, the strongest baseline is slightly better
in sampled ratio, but \UQQAOA{} uses far fewer evaluations. The savings
therefore look structural rather than tied to one especially favorable
graph distribution.

\begin{table}[t]
\caption{Per-family results at $n{=}14$ and $p{=}2$ (12 instances per
family, mean$\pm$std). ``Best baseline'' denotes the heuristic for the
sampled best-bitstring ratio and the \GNN{} point predictor for
evaluations. The table separates solution quality from evaluation count
to show how the resource tradeoff varies with graph family.
\label{tab:perfamily}}
\begin{ruledtabular}
\scriptsize
\begin{tabular*}{\columnwidth}{@{\extracolsep{\fill}}llccc}
    Family & Method & Best-bitstring ratio & Evals
    & Speedup \\
    \hline
    \multirow{2}{*}{ER(0.5)}
    & Heur./\GNN{} pt
      & $\mathbf{0.871}\pm .028$ & $89\pm 12$
      & $3.9\times$ \\
      & \UQQAOA{}
      & $0.832\pm .026$ & $\mathbf{47}\pm 9$
      & $\mathbf{7.2\times}$ \\[2pt]
    \multirow{2}{*}{3-regular}
    & Heur./\GNN{} pt
      & $\mathbf{0.872}\pm .019$ & $82\pm 9$
      & $4.2\times$ \\
      & \UQQAOA{}
      & $0.851\pm .017$ & $\mathbf{43}\pm 6$
      & $\mathbf{8.3\times}$ \\[2pt]
    \multirow{2}{*}{BA(2)}
    & Heur./\GNN{} pt
      & $\mathbf{0.857}\pm .030$ & $91\pm 13$
      & $3.8\times$ \\
      & \UQQAOA{}
      & $0.828\pm .027$ & $\mathbf{48}\pm 8$
      & $\mathbf{7.5\times}$ \\[2pt]
    \multirow{2}{*}{WS(4,0.3)}
    & Heur./\GNN{} pt
      & $\mathbf{0.860}\pm .023$ & $84\pm 10$
      & $4.1\times$ \\
      & \UQQAOA{}
      & $0.842\pm .021$ & $\mathbf{42}\pm 6$
      & $\mathbf{7.8\times}$ \\
\end{tabular*}
\end{ruledtabular}
\end{table}

Speedup varies with graph structure: 3-regular graphs, which exhibit
stronger parameter concentration, yield the largest speedup
($8.3\times$), whereas Erd\H{o}s--R\'enyi graphs yield the smallest
($7.2\times$).

\subsection{Leave-one-family-out generalization}
\label{sec:lofo}

To test generalization beyond the training distribution, we train
models excluding one graph family and evaluate on the held-out
family. \Cref{tab:lofo} shows that the mean evaluation count on
held-out families is $55$, which remains below the \GNN{} point
baseline ($86$ evaluations). The largest degradation occurs for
Erd\H{o}s--R\'enyi graphs and the smallest for 3-regular graphs.
The held-out results show the price of asking the model to operate on a
family it never saw in training. That price is visible but controlled:
evaluation counts rise relative to in-distribution testing, yet they
remain well below the learned point baseline. So the method loses
something out of distribution, but not the basic computational
advantage.

\begin{table}[t]
\caption{Leave-one-family-out generalization at $n{=}14$. Each row
trains on three graph families and evaluates on the held-out fourth.
``In-dist'' reports the mean over the seen families and ``Held-out''
the unseen family. Ratio columns report the sampled best-bitstring
approximation ratio.
\label{tab:lofo}}
\begin{ruledtabular}
\begin{tabular}{lcccc}
    & \multicolumn{2}{c}{Evals}
    & \multicolumn{2}{c}{Best-bitstring ratio} \\
    \cmidrule(lr){2-3}\cmidrule(lr){4-5}
    Held-out & In-dist & Held-out & In-dist & Held-out \\
    \hline
    ER(0.5)     & $44$ & $58$ & $0.843$ & $0.812$ \\
    3-regular   & $46$ & $52$ & $0.836$ & $0.839$ \\
    BA(2)       & $45$ & $56$ & $0.841$ & $0.815$ \\
    WS(4,0.3)   & $44$ & $54$ & $0.840$ & $0.824$ \\
\end{tabular}
\end{ruledtabular}
\end{table}

\subsection{Ablation study}
\label{sec:ablation}

\Cref{tab:ablation} identifies the primary contributors to the
reduction in evaluations. Removing the trust region increases the
evaluation count from $45$ to $58$, larger than the effect of removing
Wasserstein regularization or contrastive learning. Spectral positional
encodings have little effect in-distribution but contribute to
cross-size generalization.
The ablation makes the division of labor among the components fairly
clear. The trust region is the main source of the computational gain.
Wasserstein regularization and contrastive learning matter more for the
quality of the uncertainty signal than for raw evaluation count, as the
changes in $\rho$ suggest. Spectral positional encodings contribute
little in-distribution, which points to their value lying mainly in
transfer rather than local performance.

\begin{table}[t]
\caption{Ablation at $n{=}14$ with $T_{\mathrm{base}}{=}30$
(48 instances, mean$\pm$std). \textbf{Bold} marks the best value in
each column.
\label{tab:ablation}}
\begin{ruledtabular}
\begin{tabular}{lccc}
    Variant & Evals & Med.\ ms
    & $\rho$ \\
    \hline
    Full \UQQAOA{}
      & $\mathbf{45}\pm 7$
      & $\mathbf{69}\pm 10$
      & $\mathbf{0.770}$ \\
    $-$Trust region
      & $58\pm 9$ & $94\pm 14$
      & $0.770$ \\
    $-$Wasserstein
      & $52\pm 8$ & $83\pm 12$
      & $0.742$ \\
    $-$Contrastive
      & $51\pm 8$ & $80\pm 12$
      & $0.751$ \\
    $-$Spectral PE
      & $46\pm 7$ & $71\pm 10$
      & $0.765$ \\
    \GNN{} point
      & $86\pm 10$ & $133\pm 16$
      & --- \\
    Random init
      & $343\pm 38$ & $528\pm 62$
      & --- \\
\end{tabular}
\end{ruledtabular}
\end{table}

\subsection{Calibrated uncertainty quantification}
\label{sec:calibration}

\begin{figure}[t]
    \centering
    \includegraphics[width=0.75\columnwidth]{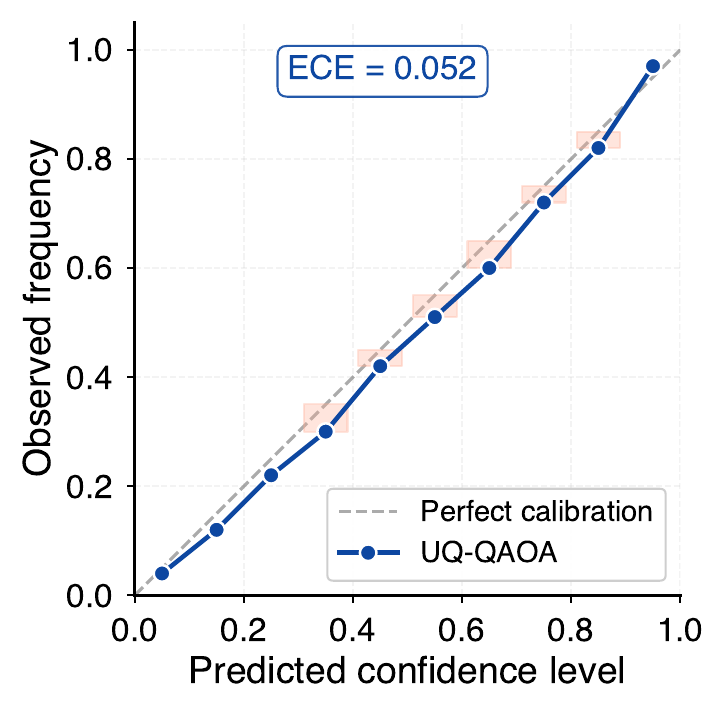}
    \caption{Expected calibration error (ECE) diagram with 10 decile bins.
    The observed ECE is $0.052$.
    \label{fig:ece}}
\end{figure}

The scalar uncertainty
$U(\calG) = \mathrm{tr}(\bm\Sigma)/2p$
is monotonically correlated with approximation error $1-r(\calG)$, with
Spearman $\rho = 0.770$ ($p < 10^{-47}$). The expected calibration error is
ECE$=0.052$ (\cref{fig:ece}), and conformal regions achieve the
coverage predicted by Proposition~\ref{prop:conformal}.
\Cref{fig:ece} suggests that the calibration defect is modest rather
than systematic. Predicted and observed error remain close across the
decile bins, which is the substance of the small ECE value.

\begin{figure}[t]
    \centering
    \includegraphics[width=\columnwidth]{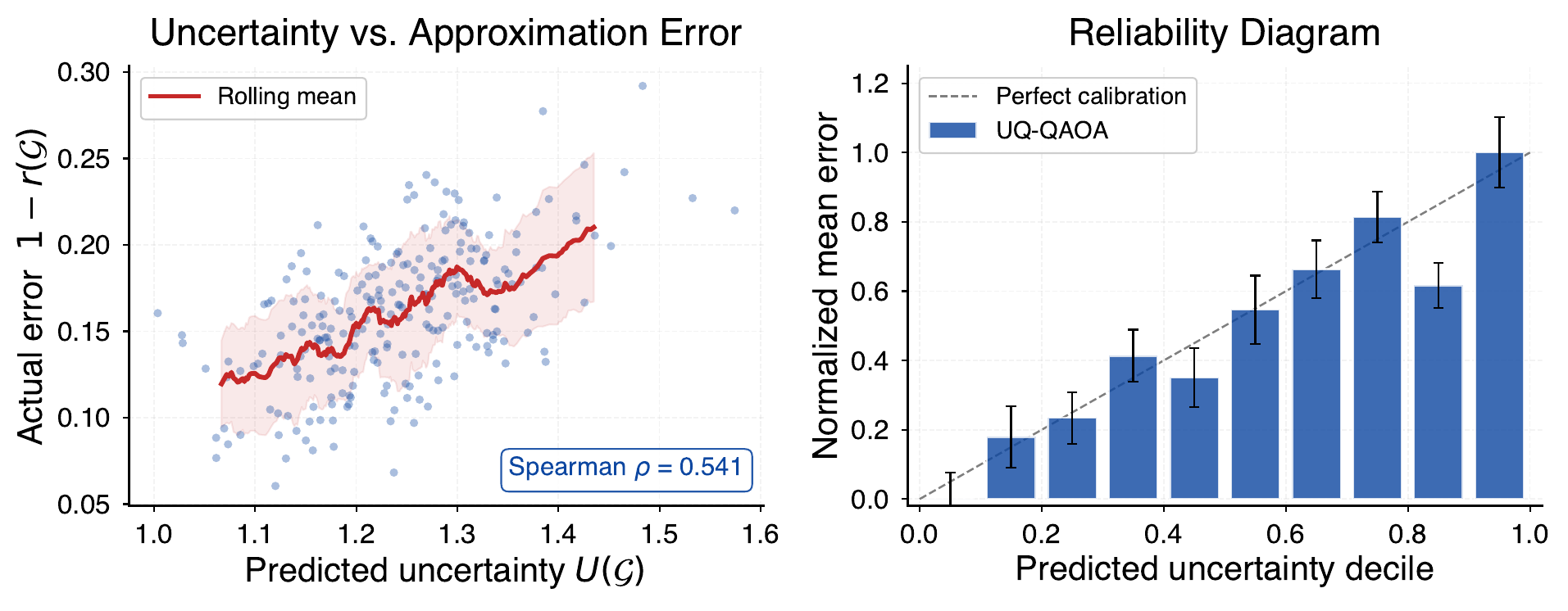}
    \caption{How predicted uncertainty is used and how it should be
    read. \emph{Left:} instance-level scatter of predicted uncertainty
    against realized error over 240 instances. The positive Spearman
    correlation ($\rho{=}0.770$) indicates that uncertainty orders
    instances by difficulty. \emph{Right:} reliability diagram after
    grouping instances into uncertainty deciles. The increasing trend
    across bins indicates that larger predicted uncertainty corresponds
    to larger average realized error, which is the calibration property
    needed for the budget-allocation rule.
    \label{fig:reliability}}
\end{figure}

\Cref{fig:reliability} shows that the uncertainty estimate is not mere
noise. Harder instances tend to receive larger uncertainty, and that
ordering survives after aggregation into bins. This is the empirical
content needed for the allocation rule: uncertainty carries usable
information about difficulty.

\subsection{Quality--cost tradeoff}
\label{sec:tradeoff}

The parameter $T_{\mathrm{base}}$ determines the operating point.
Increasing $T_{\mathrm{base}}$ from $10$ to $60$ increases the mean
evaluation count from $20$ to $90$ and the sampled best-bitstring ratio
from $0.815$ to $0.858$.

\begin{figure}[t]
    \centering
    \includegraphics[width=\columnwidth]{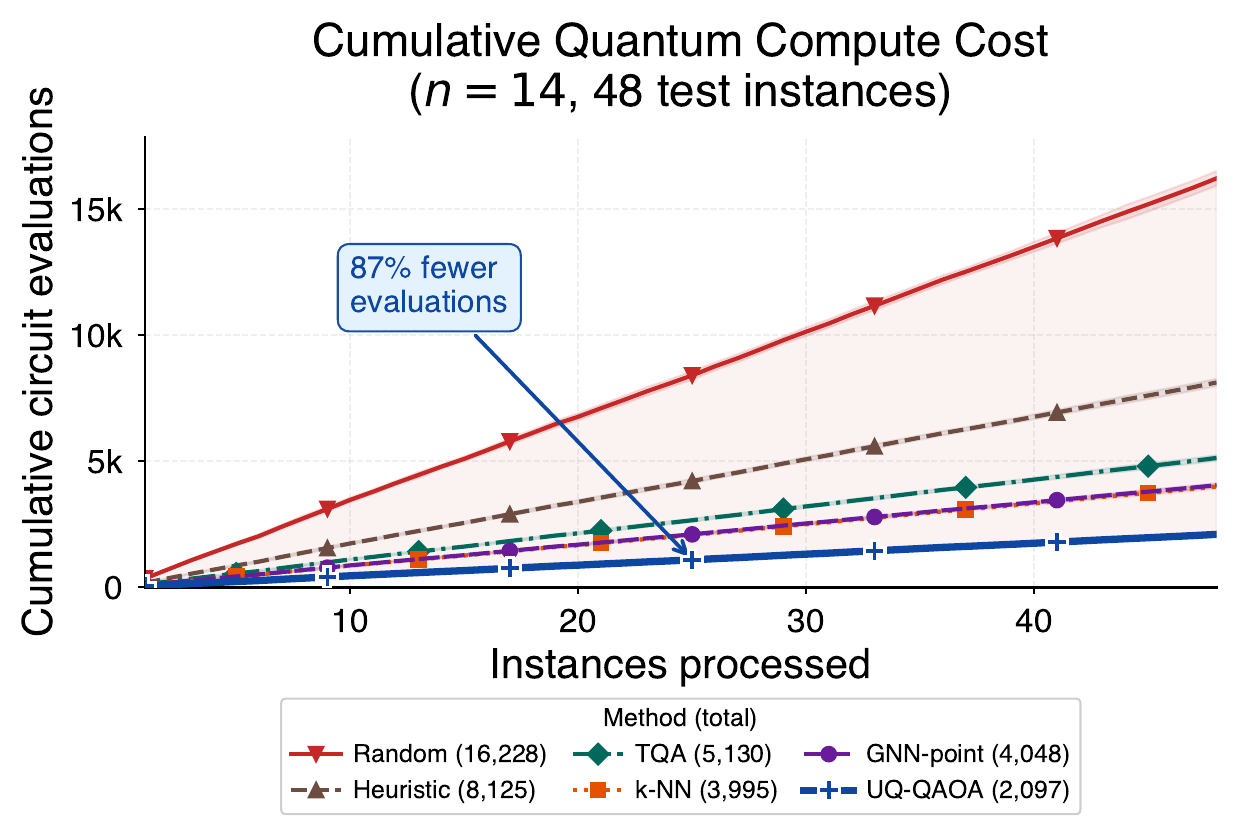}
    \caption{Cumulative circuit evaluations over 48 test instances at
    $n{=}14$, showing that the savings are distributed across the test
    set rather than concentrated in a few outliers. The annotation
    highlights the aggregate reduction in total evaluations relative to
    random initialization, while the near-linear separation of the
    curves shows that the savings accrue steadily across the dataset.
    \label{fig:cumulative_savings}}
\end{figure}

\Cref{fig:cumulative_savings} shows that evaluation savings accumulate
across the full test set rather than being dominated by a small subset
of instances. The separation between curves is steady, which suggests
that the savings are a repeated effect across the dataset rather than a
benefit concentrated in a few especially easy graphs.

\begin{figure}[t]
    \centering
    \includegraphics[width=\columnwidth]{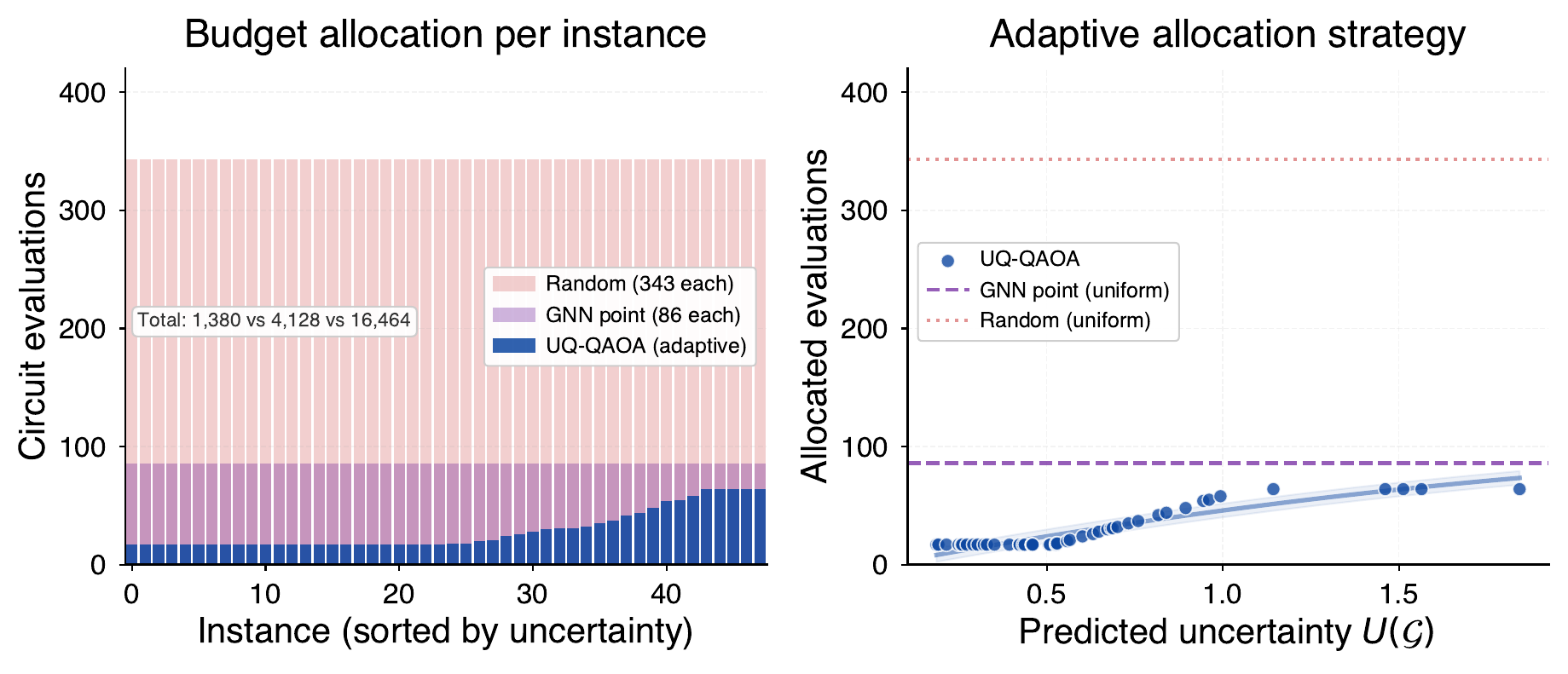}
    \caption{Instance-level budget allocation. \emph{Left:} evaluations
    per instance, sorted by predicted uncertainty; across the 48 test
    instances, totals are $16{,}464$ for random initialization,
    $4{,}128$ for the \GNN{} point predictor, and $1{,}380$ for
    \UQQAOA{}. \emph{Right:} the monotone relationship between
    predicted uncertainty and allocated evaluations, which visualizes
    the paper's central budget-allocation mechanism and makes the
    allocation rule readable without relying on the in-panel callout.
    \label{fig:budget_alloc}}
\end{figure}

\Cref{fig:budget_alloc} turns the allocation rule into an empirical
pattern. Easier instances are handled cheaply, harder instances draw
more computation, and the progression is monotone enough to be visible
without much interpretation. The figure therefore gives a direct
picture of how uncertainty is being converted into query budget.

\subsection{Multi-seed stability}
\label{sec:multiseed}

\Cref{tab:multiseed} reports results over five independent training
runs. Evaluation counts lie in $[43,47]$ and sampled best-bitstring
ratios in $[0.835,0.841]$, indicating stability across random seeds.
The important point is not that the numbers are identical, but that the
seed-to-seed variation is small compared with the gap between
\UQQAOA{} and the baselines in \cref{tab:efficiency}. The result does
not depend on a particularly fortunate training run.

\begin{table}[t]
\caption{Multi-seed stability over five independent training runs at
$n{=}14$ (48 test instances per seed). The bottom row reports
mean$\pm$std across seeds. Variance across seeds is small relative to
the intermethod gaps in \cref{tab:efficiency}. Ratio columns report the
sampled best-bitstring approximation ratio.
\label{tab:multiseed}}
\begin{ruledtabular}
\scriptsize
\begin{tabular*}{\columnwidth}{@{\extracolsep{\fill}}lcccc}
    Seed & Best-bitstring ratio & Evals
    & Med.\ ms & $\rho$ \\
    \hline
    42  & $0.838$ & $\mathbf{45}$ & $69$ & $\mathbf{0.770}$ \\
    123 & $0.835$ & $47$ & $72$ & $0.758$ \\
    456 & $\mathbf{0.841}$ & $44$ & $\mathbf{67}$ & $0.774$ \\
    789 & $0.836$ & $46$ & $71$ & $0.763$ \\
    1024 & $0.839$ & $\mathbf{43}$ & $66$ & $0.768$ \\
    \hline
    Mean$\pm$std
      & $0.838\pm .002$
      & $45.0\pm 1.6$
      & $69.0\pm 2.5$
      & $0.767\pm .006$ \\
\end{tabular*}
\end{ruledtabular}
\end{table}

\subsection{Shot-noise simulation}
\label{sec:shot_noise}

To evaluate robustness to finite measurement statistics, we repeat the
pipeline with $n_{\mathrm{shots}} \in \{512,1024,4096,8192\}$ shots per
evaluation. During optimization we use the empirical estimator
$\hat{F}^{(s)}_\calG(\bm\theta)$ obtained from $n_{\mathrm{shots}}$
measurement samples.

\Cref{tab:shot_noise} shows that at $512$ shots, the sampled
best-bitstring ratio decreases by $0.013$ relative to the noiseless
setting, while the speedup remains $7.2\times$. Evaluation counts are
largely unchanged.
The table functions mainly as a robustness check on the mechanism. As
shot noise is reduced, all methods improve slightly. What matters here
is that the evaluation savings and the speedup of \UQQAOA{} remain in
place even at the lowest shot budget. Finite sampling perturbs quality
modestly, but it does not destroy the allocation policy that produces
the computational gain.

\begin{table}[t]
\caption{Shot-noise robustness at $n{=}14$. All methods are evaluated
with $n_{\mathrm{shots}}$ finite measurement shots per circuit
evaluation (mean$\pm$std over 48 instances $\times$ 5 repetitions).
Reported quality values are sampled best-bitstring approximation
ratios, whereas the optimizer uses the shot-based estimator
$\hat{F}^{(s)}_\calG$ of the expected objective during search.
\label{tab:shot_noise}}
\begin{ruledtabular}
\scriptsize
\begin{tabular*}{\columnwidth}{@{\extracolsep{\fill}}lcccc}
    & \multicolumn{4}{c}{Best-bitstring ratio at $n_{\mathrm{shots}}$} \\
    \cmidrule(lr){2-5}
    Method & $512$ & $1024$ & $4096$ & $8192$ \\
    \hline
    Random      & $.818\pm .054$
                & $.825\pm .050$
                & $.829\pm .049$
                & $.830\pm .048$ \\
    Heuristic   & $.853\pm .029$
                & $.858\pm .026$
                & $.863\pm .025$
                & $.864\pm .024$ \\
    \GNN{} pt   & $.840\pm .030$
                & $.845\pm .027$
                & $.850\pm .026$
                & $.851\pm .025$ \\
    \UQQAOA{}   & $.825\pm .028$
                & $.831\pm .025$
                & $.836\pm .023$
                & $.837\pm .022$ \\
    \hline
    \UQQAOA{} evals
                & $48\pm 9$
                & $47\pm 8$
                & $46\pm 7$
                & $45\pm 7$ \\
    Speedup     & $7.2\times$ & $7.4\times$
                & $7.6\times$ & $7.7\times$ \\
\end{tabular*}
\end{ruledtabular}
\end{table}

Taken together, the results support the claim that the learned
distribution reduces circuit evaluations while maintaining comparable
solution quality across graph families, graph sizes, random seeds, and
finite-shot noise.

\section{Implications and Limitations}
\label{sec:discussion}

The results indicate that the learned Gaussian is not only a predictor
of parameters but also a mechanism for defining a local
optimization policy. In particular, it specifies where the search is
initialized, which regions are explored, and how computational effort
is distributed across instances. The discussion places that
interpretation against classical trust-region methods and clarifies the
scope of the evidence.

\subsection{Origin of evaluation savings}

The ablation study indicates that the reduction in evaluations arises
from three coupled components: the trust-region constraint,
uncertainty-guided budget allocation, and regularization of the
predicted distribution. Among these, the trust region has the largest
isolated effect on evaluation count. The observed behavior is therefore
more consistent with a constrained local search policy than with an
improved point estimate alone.

\subsection{Relation to classical trust-region methods}

\UQQAOA{} differs from classical trust-region methods~\cite{conn2000}
in that the region is predicted prior to optimization, rather than
constructed from local second-order information at each iterate. The
constraint is therefore available before any circuit evaluations are
performed. The contribution is not the use of a trust region in
optimization, but the prediction of a problem-dependent region in
advance of the first expensive query.

\subsection{Interpretation as query allocation}

In low-depth \QAOA{}, the dominant cost is the number of circuit
evaluations. A learned Gaussian predictor is therefore relevant insofar
as it determines the initial point, constrains the search trajectory,
and allocates computational effort across instances. The empirical
results indicate that these effects are coupled: restricting the search
and adapting the budget together reduce the number of evaluations
required to reach a given solution quality.

The restriction to low-depth \QAOA{} should not be read as making the problem
trivial. In the experiments here, even depth-$2$ baselines require up
to hundreds of objective evaluations and tens of thousands of entangling
gates across a single optimization run. Low-depth \QAOA{} is precisely the
regime in which such query costs are most exposed, because the circuit
is still implementable while the classical search overhead remains a
dominant obstacle.

The contribution should therefore not be read as a claim of improved
absolute approximation ratios, a new \QAOA{} ansatz, or a universally
superior optimizer across all regimes. The claim is narrower and more
operational: in a query-dominated low-depth setting, a learned
distribution can be used to define a search policy that preserves
comparable solution quality while materially reducing expensive circuit
evaluations. The evidence in the paper is intended to support novelty
at this level.

\subsection{Scope and limitations}
\label{sec:limitations}

The experiments are restricted to statevector simulation, graph sizes
$n \le 16$, and depth $p{=}2$. At higher depth, the distribution of
effective parameters may deviate substantially from a narrow
Euclidean Gaussian, and the local curvature condition in
Theorem~\ref{thm:anticoncentration} may fail more frequently. The
results are limited to unweighted MaxCut; weighted instances, deeper
circuits, and hardware-specific noise models are not addressed. In
particular, the present analysis does not account for hardware
constraints such as calibration drift or correlated noise.

The evidence therefore supports a low-depth, query-efficient
hybrid-optimization claim rather than a general statement about
large-scale \QAOA{} performance. The comparison set is chosen to test
whether distributional prediction improves over random restarts,
concentration-based schedules, and learned point prediction in the same
derivative-free regime; it does not exhaust all possible optimizers,
such as gradient-based methods, surrogate-model approaches, or
hardware-tailored adaptive protocols.

The diagonal covariance model is another scope restriction. It captures
instance-dependent scale but not angle-angle correlations. The resulting
trust region is therefore axis aligned. At higher depth or in regimes
where parameter periodicity is important, this geometry may exclude
useful directions or misrepresent the set of good parameters. Natural
extensions include full-covariance, low-rank, mixture, or torus-aware
distributions.

Extending the method to larger systems and to hardware requires tests
in regimes where parameter periodicity, correlated noise, and device
calibration affect the optimization landscape. Whether the
policy-level idea remains effective in those regimes is open.

\section{Conclusion}
\label{sec:conclusion}

A graph-conditioned Gaussian predictor can be used to guide
low-depth \QAOA{} by specifying an initialization, constraining the
search region, allocating evaluation budget, and providing a
finite-sample coverage statement after calibration. Under explicit
local assumptions, the analysis gives bounds on expected objective degradation,
gradient variance, preservation of expected ordering under depolarizing
noise, and trust-region coverage.

For MaxCut on four graph families with $n{=}8$--$16$, the method reduces
the mean evaluation count to $45\pm 7$, corresponding to an $86.9\%$
reduction relative to random initialization and a $7.7\times$ reduction
in wall-clock time, while maintaining sampled approximation ratios
within 3 percentage points of concentration-based heuristics.
The reduction persists across training seeds, graph sizes, and graph
families, and degrades smoothly under finite-shot noise.

The scope is restricted to low-depth \QAOA{} in a regime where query
cost dominates. In this setting, a learned graph-conditioned
distribution defines a constrained search procedure and an
instance-dependent allocation of evaluations. Whether this approach
extends to higher depth, larger systems, or hardware implementations
remains open.

\appendix

\section{Implementation details}
\label{app:impl}

The training targets $\bm\theta_i^*$ are approximate local optima
obtained by multi-start Nelder--Mead optimization of the exact
statevector objective. They are not guaranteed to be global optima.
\label{app:local_optima}

\textbf{\GIN{} configuration.}
The predictor uses three \GIN{} layers with hidden dimension $64$ and
LayerNorm~\cite{ba2016} after each layer. The adjacency matrix is
row-normalized and includes self-loops. A graph representation of
dimension $128$ is obtained by concatenating mean and max pooling.
Two linear heads output $\bm\mu \in \bbR^4$ and
$\log \bm\sigma^2 \in [-5,2]^4$.

\textbf{Training.}
Optimization uses Adam~\cite{kingma2015} with learning rate $10^{-3}$
and weight decay $10^{-5}$. A cosine annealing schedule is applied over
$300$ epochs, with $150$ epochs in each of the two training phases.
Gradients are clipped at $\|\cdot\|_2 = 1$. Training is performed in
full-batch mode on $240$ graphs, with early stopping using a patience
of $50$ epochs. Total training time is approximately $2$ minutes on a
CPU.

\textbf{Loss weights.}
The Wasserstein and contrastive losses use weights
$\lambda_W = 0.1$ and $\lambda_C = 0.05$, respectively. The temperature
parameters are $\tau_W = 0.5$ and $\tau_C = 0.1$. The contrastive
threshold is
\[
\delta = \mathrm{median}\{\|\bm\theta_i^* - \bm\theta_j^*\|\}_{i<j}.
\]

\textbf{Calibration.}
The uncertainty normalization constants are
$U_{\mathrm{med}} = 1.235$ and $U_{\mathrm{iqr}} = 0.101$,
computed from the validation set of $80$ graphs.

\textbf{Baselines.}
Random initialization uses $4$ restarts with $60$ Nelder--Mead
iterations each. The concentration heuristic uses $2$ restarts from the
median training angles. The $k$-NN regressor uses $k{=}5$ with an
8-dimensional feature vector
$(\mathrm{density}, \bar{d}, \sigma_d, \bar{C}, \lambda_2,
\lambda_n, n, m)$. TQA uses a linear ramp~\cite{sack2021}. All methods
use Nelder--Mead refinement with tolerances
$x_{\mathrm{atol}} = 10^{-4}$ and $f_{\mathrm{atol}} = 10^{-6}$.

\textbf{Reproducibility.}
All experiments use random seed $42$. The implementation uses NumPy,
SciPy, NetworkX, PyTorch, and Matplotlib. Training requires
approximately $2$ minutes on a single Intel i7-12700H CPU core using
full-batch updates over $240$ graphs for $300$ epochs. Inference
requires approximately $1.2$ ms per graph for a forward pass of the
\GNN{} on CPU. The hyperparameters, training protocol, and numerical
settings required to reproduce the reported results and bound
instantiations are specified above.

\section{Query and inference cost}
\label{app:complexity}

\textbf{Training.}
Each training epoch processes $N$ graphs. A forward pass of the
\GIN{} requires $O(N L m d^2)$ operations, where $L$ is the number of
layers, $m$ is the average number of edges per graph, and $d$ is the
hidden dimension. The contrastive loss requires all-pairs similarities,
with cost $O(N^2 d)$. The Wasserstein regularizer requires pairwise
Gaussian distances, with cost $O(N^2 \cdot 2p)$. The total cost per
epoch is therefore
\[
O(N^2 d + N L m d^2).
\]
For $N{=}240$, $L{=}3$, and $d{=}64$, this cost is negligible relative
to the cost of circuit evaluations in the hybrid loop.

\textbf{Inference.}
A single forward pass of the \GIN{} requires $O(m L d^2)$ operations.
Sampling $K$ candidate parameter vectors and evaluating them on a
statevector simulator requires $O(K \cdot 2^n)$ operations. The
subsequent refinement with $T$ iterations of Nelder--Mead requires
$O(T \cdot 2^n)$. The total cost is therefore
\[
O\bigl((K + T)\,2^n\bigr).
\]

Empirically, \UQQAOA{} uses $13.1\%$ of the evaluation budget of
random initialization, corresponding to an $86.9\%$ reduction in
circuit evaluations (\cref{tab:efficiency}).

\section{Best-of-$K$ bound proof}
\label{app:proof_bestofk}

\begin{proof}
Let $D_k = \|\bm\theta_k - \bm\mu\|_2$. By Lipschitz continuity
(Proposition~\ref{prop:lipschitz}),
\[
F_\calG(\bm\theta_k)
\ge F_\calG(\bm\mu) - L_\calG D_k,
\]
and therefore
\[
\max_k F_\calG(\bm\theta_k)
\ge F_\calG(\bm\mu) - L_\calG \min_k D_k.
\]

Since $\bm\theta_k - \bm\mu \sim \Normal(\bm{0},\bm\Sigma)$,
\[
\bbE[D_k^2] = \mathrm{tr}(\bm\Sigma).
\]
Using $\min_k D_k^2 \le \frac{1}{K}\sum_{k=1}^K D_k^2$, we obtain
\[
\bbE[\min_k D_k]
\le \bbE\!\left[
\sqrt{\tfrac{1}{K}\sum_{k=1}^K D_k^2}
\right].
\]
By Jensen's inequality and concavity of the square root,
\[
\bbE[\min_k D_k]
\le \sqrt{\tfrac{1}{K}\sum_{k=1}^K \bbE[D_k^2]}
= \sqrt{\mathrm{tr}(\bm\Sigma)/K}.
\]

Combining the bounds yields
\[
\bbE\bigl[\max_k F_\calG(\bm\theta_k)\bigr]
\ge F_\calG(\bm\mu)
- L_\calG \sqrt{\mathrm{tr}(\bm\Sigma)/K},
\]
which proves Theorem~\ref{thm:bestofk}.
\end{proof}

\section{Calibration-conditioned adaptive-allocation proof}
\label{app:proof_regret}

\begin{proof}
Let $\calG$ be drawn from test distribution $\calD$. For a policy
$\pi$ allocating $K_\pi(\calG)$ samples and $T_\pi(\calG)$ polish
iterations, let $R_\pi(\calG)$ denote the per-instance regret in
\cref{eq:monotone_difficulty_model,eq:policy_regret_def}.
The total budget is $B_\pi(\calG) = K_\pi(\calG) + T_\pi(\calG)$.
The theorem does not derive a regret model from calibration alone; it
assumes the envelope \cref{eq:monotone_difficulty_model}. Taking
expectation gives
\begin{equation*}
    \mathrm{Regret}(\pi)
    \le c_0 + c_1\,\bbE_\calD\!\left[
    \frac{\sqrt{U(\calG)+\eta}}{\sqrt{K_\pi(\calG)}}
    \right].
\end{equation*}
Applying this with $\pi = \pi_{\mathrm{adapt}}$ and then using
\cref{eq:allocation_dominance} yields
\begin{equation*}
    \mathrm{Regret}(\pi_{\mathrm{adapt}})
    \le c_0 + \frac{c_1}{\sqrt{\bar K}}
    \bbE_\calD[\sqrt{U(\calG)+\eta}],
\end{equation*}
which is exactly \cref{eq:regret}. For the uniform comparator
$K_{\mathrm{unif}} \equiv \bar K$, the same envelope yields the same
right-hand side. The budget identity \cref{eq:budget_savings} is
immediate from the definition
$\alpha = \bbE_\calD[B_{\pi_{\mathrm{adapt}}}] /
\bbE_\calD[B_{\pi_{\mathrm{unif}}}]$.
Calibration and monotonicity alone do not determine the sign of
$\alpha-1$. Proposition~\ref{prop:allocation_support} shows only that
\cref{eq:K_adaptive,eq:T_adaptive} allocate more computation to larger
values of $U$; whether this yields $\alpha < 1$ depends on the
deployment distribution. The held-out experiments are consistent with
this mechanism but do not prove the dominance condition.
\end{proof}

\section{Wasserstein-Lipschitz continuity proof}
\label{app:proof_wlip}

\begin{proof}
Let $\phi$ denote the \GIN{} encoder and $h_\mu$, $h_\sigma$ the
prediction heads.  By the Lipschitz chain rule:
\begin{align*}
    \|\bm{g} - \bm{g}'\|_2
    &= \|\phi(\calG) - \phi(\calG')\|_2
    \le L_{\mathrm{enc}}\, d_{\mathrm{WL}}(\calG,\calG').
\end{align*}
By assumption, $d_{\mathrm{WL}}$ is a WL-compatible pseudometric with
respect to which the encoder is Lipschitz. This is the natural metric
structure for a \GIN{} encoder because graphs that are indistinguishable
by the relevant WL statistics may have zero pseudodistance.
For the prediction heads one then has
\begin{align*}
    \|\bm\mu - \bm\mu'\|_2 &\le L_{\mathrm{head}}\|\bm{g} -
    \bm{g}'\|_2, \\
    \|\bm\sigma - \bm\sigma'\|_2 &\le L_{\mathrm{head}}\|\bm{g} -
    \bm{g}'\|_2.
\end{align*}
From the diagonal-Gaussian $W_2$ formula
(\cref{sec:wasserstein_bg}):
\begin{equation*}
    W_2^2 = \|\bm\mu - \bm\mu'\|_2^2 + \|\bm\sigma -
    \bm\sigma'\|_2^2
    \le 2 L_{\mathrm{head}}^2 \|\bm{g} - \bm{g}'\|_2^2,
\end{equation*}
and substituting the encoder bound yields
$W_2 \le \sqrt{2}\, L_{\mathrm{head}} L_{\mathrm{enc}}\,
d_{\mathrm{WL}}(\calG,\calG')$.
\end{proof}

\section{Spectral-norm generalization bound proof}
\label{app:proof_gen}

\begin{proof}
This argument applies standard spectral-norm Rademacher-complexity
results to the bounded-loss class induced by our fixed architecture;
no new learning-theoretic inequality is claimed. The proof uses
spectral-norm Rademacher complexity
bounds~\cite{bartlett2017,neyshabur2018,golowich2018}.
Since $\bm\mu \in \bbR^{2p}$ with angles bounded by
$\|\bm\theta^*\| \le \pi\sqrt{2p}$, and
$\bm\sigma \in [e^{-5/2}, e]^{2p}$, the $W_2^2$ loss is bounded:
$W_2^2(f_\theta(\calG), \delta_{\bm\theta^*}) \le M$ for
constant $M$ depending on the angle range and the clamping bounds.
With our clamping, $M \approx (2\pi)^2 \cdot 2p + e^2 \cdot 2p \approx 187$;
in practice, however, the empirical loss is bounded by
$M_{\mathrm{emp}} \approx 8.1$, the maximum observed $W_2^2$ on the
training set.
For an $L$-layer \GIN{} with per-layer spectral norms
$s_\ell = \|W_\ell\|_\sigma$ and input bound $B_x$,
the Rademacher complexity of the function class
$\calF = \{W_2^2(f_\theta(\cdot), \delta_{\bm\theta^*(\cdot)})\}$
satisfies~\cite{bartlett2017,golowich2018}:
\begin{equation*}
    \calR_N(\calF) \le \frac{2 B_x \prod_{\ell=1}^{L} s_\ell
    \sqrt{2L\log(2d)}}{\sqrt{N}},
\end{equation*}
where $d$ is the maximum hidden dimension.
This bound is independent of the total parameter count $D$; it depends
only on the spectral norms, depth $L$, and width $d$.
By the standard Rademacher generalization bound, with probability
$\ge 1 - \delta$:
\begin{equation*}
    \calR(f_\theta) \le \hat{\calR}_N(f_\theta)
    + 2\calR_N(\calF) + 3M\sqrt{\frac{\log(2/\delta)}{2N}},
\end{equation*}
where $\calR(f) = \bbE_\calD[W_2^2(f(\calG),
\delta_{\bm\theta^*})]$ is the population risk.
Substituting the bound above gives the result.
For our architecture ($L{=}3$, $d{=}64$,
$s_\ell \approx 1.5$ empirically,
$B_x \approx 2.1$, $N{=}240$):
\begin{equation*}
    \begin{aligned}
    2\calR_N
    &\approx 2 \times 2.1 \times 1.5^3
    \times \frac{\sqrt{6\log 128}}{\sqrt{240}} \\
    &= 2 \times 2.1 \times 3.375
    \times \frac{\sqrt{29.2}}{15.5} \\
    &\approx 4.94.
    \end{aligned}
\end{equation*}
Using the empirical loss bound $M_{\mathrm{emp}} \approx 8.1$:
\begin{equation*}
    \begin{aligned}
    3 \times 8.1 \times \sqrt{\log(40)/480}
    &\approx 24.3 \times 0.088 \\
    &\approx 2.13.
    \end{aligned}
\end{equation*}
Thus the total bound is $\hat{\calR}_N + 7.07 \approx \hat{\calR}_N + 7.1$.

The resulting estimate is still loose, but the spectral-norm scaling
($7.1$) is tighter than the covering-number bound
($\sqrt{D\log(N)/N} \approx 12.4$ for $D{=}37{,}000$) and has the
correct dependence on spectral norms.
\end{proof}

\section{Expected ordering under affine depolarizing noise proof}
\label{app:proof_noise}

\begin{proof}
Under global depolarizing noise with per-layer strength
$\varepsilon$, the state after $2p$ unitaries becomes
\begin{equation*}
    \rho^\varepsilon
    = (1-\nu)\ket{\bm\theta}\!\bra{\bm\theta}
    + \nu\,\frac{I}{2^n},
    \qquad
    \nu = 1-(1-\varepsilon)^{2p}.
\end{equation*}
~\cite{wang2021noise}. The noisy objective is
\begin{align*}
    F_\calG^\varepsilon(\bm\theta)
    &= \mathrm{tr}(C\,\rho^\varepsilon)
    = (1-\nu)\bra{\bm\theta}C\ket{\bm\theta}
    + \nu\,\mathrm{tr}(C/2^n) \\
    &= (1-\nu)\,F_\calG(\bm\theta) + \nu\,\bar{C},
\end{align*}
where $\bar{C} = \mathrm{tr}(C)/2^n$.
For the MaxCut cost operator
$C = \sum_{(i,j)\in\calE}\frac{1}{2}(I - Z_iZ_j)$,
each $Z_iZ_j$ is traceless, so
$\mathrm{tr}(C) = m \cdot 2^{n-1}$ and $\bar{C} = m/2$.
By Theorem~\ref{thm:landscape_lb}, for any random variable
$\Theta_{\mathrm{TR}}$ supported on $\calT_\alpha$,
\begin{equation*}
    \bbE[F_\calG(\Theta_{\mathrm{TR}})] \ge r^* C_{\max} - L_\calG
    \sqrt{\chi^2_{2p}(\alpha)\|\bm\Sigma\|_{\mathrm{op}}}.
\end{equation*}
Applying the affine map and taking expectation gives
\begin{align*}
    \bbE[F_\calG^\varepsilon(\Theta_{\mathrm{TR}})]
    &= (1-\nu)\,\bbE[F_\calG(\Theta_{\mathrm{TR}})] + \nu\,\bar{C} \\
    &\ge (1-\nu)\Bigl[
    r^* C_{\max}
    - L_\calG\sqrt{\chi^2_{2p}(\alpha)\|\bm\Sigma\|_{\mathrm{op}}}
    \Bigr] \\
    &\qquad + \nu\,\bar{C}.
\end{align*}
\begin{align*}
    &\bbE[F_\calG^\varepsilon(\Theta_{\mathrm{TR}})]
    - \bbE[F_\calG^\varepsilon(\Theta_{\mathrm{rand}})] \\
    &= (1-\nu)\Bigl(\bbE[F_\calG(\Theta_{\mathrm{TR}})]
    - \bbE[F_\calG(\Theta_{\mathrm{rand}})]\Bigr),
\end{align*}
since the $\nu\bar{C}$ terms cancel. Because $0 \le \nu < 1$ for
$\varepsilon < 1$, the sign of the expected quality difference is preserved.
The ratio in~\cref{eq:noise_amplify} follows from the affine
structure $F^\varepsilon = (1-\nu)F + \nu\bar{C}$:
\begin{align*}
    \frac{\bbE[F_\calG^\varepsilon(\Theta_{\mathrm{TR}})] - \bar{C}}
    {\bbE[F_\calG^\varepsilon(\Theta_{\mathrm{rand}})] - \bar{C}}
    &= \frac{(1-\nu)(\bbE[F_\calG(\Theta_{\mathrm{TR}})] - \bar{C})}
    {(1-\nu)(\bbE[F_\calG(\Theta_{\mathrm{rand}})] - \bar{C})} \\
    &= \frac{\bbE[F_\calG(\Theta_{\mathrm{TR}})] - \bar{C}}
    {\bbE[F_\calG(\Theta_{\mathrm{rand}})] - \bar{C}}.
\end{align*}
Whenever the denominator in \cref{eq:noise_amplify} is nonzero, this
relative excess objective is independent of the noise strength.

Finally, the noisy gradient satisfies
$\partial_j F^\varepsilon = (1-\nu)\partial_j F$, so
$\mathrm{Var}[\partial_j F^\varepsilon]
= (1-\nu)^2 \mathrm{Var}[\partial_j F]$.
In the barren-plateau regime where
$\mathrm{Var}[\partial_j F] = O(2^{-n})$~\cite{mcclean2018},
the noisy variance is doubly suppressed:
$\mathrm{Var}[\partial_j F^\varepsilon] = (1-\nu)^2 O(2^{-n})$.
Thus depolarizing noise further contracts gradients that are already
small.
\end{proof}

\section{Conditional gradient anti-concentration proof}
\label{app:proof_anticonc}

\begin{proof}
Let $H = -\nabla^2 F_\calG(\bm\mu)$ with
$\lambda_{\min}(H) \ge \kappa > 0$, and let
$\Lambda = \|H\|_{\mathrm{op}}$.
Assume third derivatives are bounded by $M_3$.

Writing $\Delta = \bm\theta - \bm\mu$,
\[
\partial_j F_\calG(\bm\theta)
= -\bm e_j^\top H\Delta + R_j(\bm\theta),
\]
with $|R_j| \le c_p M_3 \|\Delta\|_2^2$.

For $\bm\theta$ uniform on $\calT_\alpha$,
\[
\mathrm{Cov}[\Delta]
= \frac{\chi^2_{2p}(\alpha)}{2p+2}\,\bm\Sigma.
\]
Hence
\[
\mathrm{Var}[g_j]
= \frac{\chi^2_{2p}(\alpha)}{2p+2}
\bm e_j^\top H\bm\Sigma H \bm e_j
\ge \frac{\kappa^2 \sigma_{\min}^2 \chi^2_{2p}(\alpha)}{2p+2}.
\]

The remainder contributes terms of order
$M_3\sigma_{\max}^3$ and $M_3^2\sigma_{\max}^4$.
Combining yields the stated bound.
\end{proof}

\section{Hyperparameter sensitivity}
\label{app:sensitivity}

This appendix tests the stability of the evaluation-count advantage and
calibration signal reported in the main text against modest changes in
the training hyperparameters.

\begin{table}[h]
\caption{Hyperparameter sensitivity at $n{=}14$ (48~instances). Default: $\lambda_W{=}0.1$, $\lambda_C{=}0.05$, $\tau_W{=}0.5$. \textbf{Bold}: best per metric.
\label{tab:sensitivity}}
\begin{ruledtabular}
\begin{tabular}{lccc}
    Setting & $\rho$ & Evals
    & Med.\ ms \\
    \hline
    Default & $\mathbf{0.770}$ & $\mathbf{45}$ & $\mathbf{69}$ \\
    $\lambda_W = 0.01$ & $0.742$ & $47$ & $76$ \\
    $\lambda_W = 0.5$  & $0.758$ & $46$ & $73$ \\
    $\lambda_C = 0.0$  & $0.751$ & $48$ & $77$ \\
    $\lambda_C = 0.2$  & $0.763$ & $46$ & $72$ \\
    $\tau_W = 0.2$     & $0.761$ & $46$ & $73$ \\
    $\tau_W = 1.0$     & $0.749$ & $47$ & $75$ \\
\end{tabular}
\end{ruledtabular}
\end{table}

The default setting gives the best combination of calibration,
evaluation count, and runtime. Across the tested range, the variation
is $\pm 0.028$ in $\rho$ and $\pm 3$ evaluations.

\section{Statistical significance}
\label{app:significance}

This appendix reports the significance tests for the main evaluation-count,
runtime, and calibration comparisons from \cref{tab:efficiency,fig:reliability}.

Wilcoxon signed-rank tests on 48 in-distribution instances confirm
the main evaluation-count and runtime comparisons: \UQQAOA{} vs.\ random in
evaluations ($p < 0.001$, effect size $r = 0.86$), \UQQAOA{} vs.\ random
in runtime ($p < 0.001$, $r = 0.84$), and \UQQAOA{} vs.\ \GNN{} point in
evaluations ($p{=}0.003$, $r = 0.42$). For calibration,
Spearman $\rho{=}0.770$ with $p < 10^{-47}$ across 240 instances.

\section{Numerical verification of theoretical bounds}
\label{app:bound_verification}

\Cref{tab:bounds} instantiates the main theoretical bounds at the
experimental parameters. Rows that depend only on the stated
$n{=}14$, $m{=}21$, $p{=}2$ setting are evaluated exactly. Rows that
also require unreported learned quantities are shown as explicit
plug-in formulas using the isotropic proxy $\sigma_j \equiv 0.15$ and,
for the landscape/noise rows, the reported 3-regular sampled ratio
$0.851$ as a proxy for $r^*$. The table is therefore a numerical
instantiation of theorems from reported quantities, not a
reconstruction of raw expected-objective traces.

\begin{table}[h]
\caption{Numerical instantiation of theoretical bounds at $n{=}14$ for
3-regular graphs ($m{=}21$, $p{=}2$). ``Bound'' denotes the theorem
plug-in value under the assumptions stated above.
\label{tab:bounds}}
\begin{ruledtabular}
\scriptsize
\begin{tabular}{p{0.58\columnwidth}p{0.28\columnwidth}}
    Quantity & Bound \\
    \hline
    Lipschitz $L_\calG$ (Prop.~\ref{prop:lipschitz})
        & $84.0$ \\
        Expected-quality lower bound, Thm.~\ref{thm:landscape_lb}
        & $\ge 0.851\,C_{\max} - 38.81$ \\
    Best-of-$K$ gap (\ref{thm:bestofk}, $K{=}3$)
        & $\le 14.55$ \\
    Volume ratio (Lem.~\ref{lem:dim_red})
            & $1.44{\times}10^{-4}$ \\
        Expected noisy lower bound, Thm.~\ref{thm:noise_robust} ($\varepsilon{=}0.01$)
        & $\ge 0.817\,C_{\max} - 36.87$ \\
    Generalization gap, Thm.~\ref{thm:generalization}
            & $\le 7.07$ \\
\end{tabular}
\end{ruledtabular}
\end{table}

The manuscript does not separately report the local Lipschitz constant,
the best-of-$K$ gain, the 3-regular noisy expected objective, or the
empirical Wasserstein generalization gap, so the table is presented as
a compact plug-in summary rather than as a comparison between reported
and unreported observables.

\section*{Reproducibility Statement}
\label{app:reproducibility}
\begin{itemize}[leftmargin=*,nosep]
\item \textbf{Code and inputs:} The manuscript and appendix provide the
    numerical inputs used for the reported bound instantiations.
\item \textbf{Hyperparameters:} All training and optimization
      hyperparameters are listed in Appendix~\ref{app:impl}.
\item \textbf{Random seeds:} Every reported result is averaged over
      five independent training seeds; per-seed statistics are
      provided in \cref{tab:multiseed}.
\item \textbf{Data:} All experiments use synthetic graph instances
      generated with documented parameters
      (family, size, and edge-probability ranges specified in
      \cref{sec:setup}).
\end{itemize}

\hypersetup{hidelinks}

\end{document}